\pdfoutput=1

\documentclass[11pt]{article}

\usepackage{acl}

\usepackage{times}
\usepackage{latexsym}
\usepackage{amsmath}
\usepackage{comment}
\usepackage[T1]{fontenc}

\usepackage[utf8]{inputenc}
\usepackage{graphicx}
\usepackage{paralist, tabularx}
\usepackage{microtype}
\usepackage{makecell}
\usepackage{multirow}
\usepackage{multicol}
\usepackage{tabu}
\usepackage{booktabs}
\usepackage{hyperref}
\usepackage{subcaption}
\usepackage{inconsolata}
\definecolor{lightblue}{RGB}{173, 216, 230}
\usepackage{mathtools,nicefrac}
\usepackage{xspace}

\newcommand{\sca}{\textsc{Self-Contra}\xspace}
\usepackage[normalem]{ulem}

\title{Self-Contradictory Reasoning Evaluation and Detection}

\author{Ziyi Liu$^{1}$ \hspace{3mm} Soumya Sanyal$^{1}$\hspace{3mm} Isabelle Lee$^{1}$ \hspace{3mm} Yongkang Du$^{1}$\hspace{3mm} \\ \textbf{Rahul Gupta}$^{2}$\hspace{3mm} \textbf{Yang Liu}$^{2}$\hspace{3mm} \textbf{Jieyu Zhao}$^{1}$ \\
$^{1}$University of Southern California \hspace{3mm} $^{2}$Amazon AGI\hspace{3mm}\\
\small{\texttt{\{zliu2803, soumyasa, gunheele, duyongka, jieyuz\}@usc.edu}}\\
\small{\texttt{\{gupra, yangliud\}@amazon.com}}\\
}

\begin{document}
\maketitle
\begin{abstract}
In a plethora of recent work, large language models (LLMs) demonstrated impressive reasoning ability, but many proposed downstream reasoning tasks only focus on final answers. Two fundamental questions persist: 1) how consistent is the reasoning, and 2) can models detect unreliable reasoning? In this paper, we investigate self-contradictory (\sca) reasoning, where the model reasoning does not support its answers. To answer 1), we define and assess the \sca 
rate across three datasets and delve into finer-grained categories of \sca reasoning. We find that LLMs often contradict themselves in reasoning tasks involving contextual information understanding or commonsense. 
The model may generate correct answers by taking shortcuts in reasoning or overlooking contextual evidence, leading to compromised reasoning. For 2), we task the state-of-the-art model GPT-4
with identifying \sca reasoning and finer-grained fallacies. We find that finer-grained categories enhanced detection can improve GPT-4's ability to detect \sca. However, it is only able to detect \sca with a 52.2\% F1 score, much lower compared to 66.7\% for humans. Our results indicate that current LLMs lack the robustness necessary for reliable reasoning and we emphasize the urgent need for establishing best practices in comprehensive reasoning evaluations beyond pure performance-based metrics.\footnote{The code and dataset are available at \href{https://github.com/uscnlp-lime/Self-Contradictory}{https://github.com/uscnlp-lime/Self-Contradictory}.}

\end{abstract}

\section{Introduction}

\begin{figure}[!th]
    \centering
    \includegraphics[width=1\columnwidth]{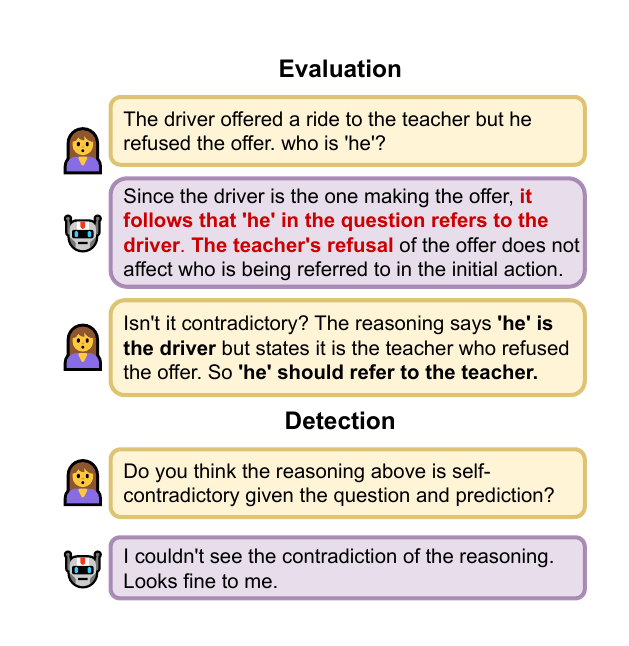}
    \caption{An example for self-contradictory reasoning and detection by LLMs. LLMs fail to generate consistent reasoning and are poor at detecting the self-contradiction.} 
    \label{fig:intro}
\end{figure}

Large language models (LLMs) have shown impressive performance in many NLP tasks, such as question answering \citep{wang2022pinto}, and math reasoning \citep{wang2022self,wei2022chain,lyu2023faithful,kojima2022large}. 
LLMs can achieve high accuracy on reasoning datasets such as CommonSenseQA \citep{bauer2018commonsense} with carefully designed prompts. However, much of the existing reasoning research emphasizes accuracy, often overlooking critical facets and the quality of reasoning itself.
In fact, a correct prediction does not necessarily reflect sound reasoning as a model could predict based on spurious correlations~\cite{mccoy2019right}. To build trustworthy models, it is crucial to maintain a coherent and consistent logical connection between a model's predictions and its reasoning. The lack of trustworthiness can hurt human confidence in LLMs \cite{liu2023trustworthy}.
\begin{figure*}[t]
    \centering
    \includegraphics[width=2\columnwidth]{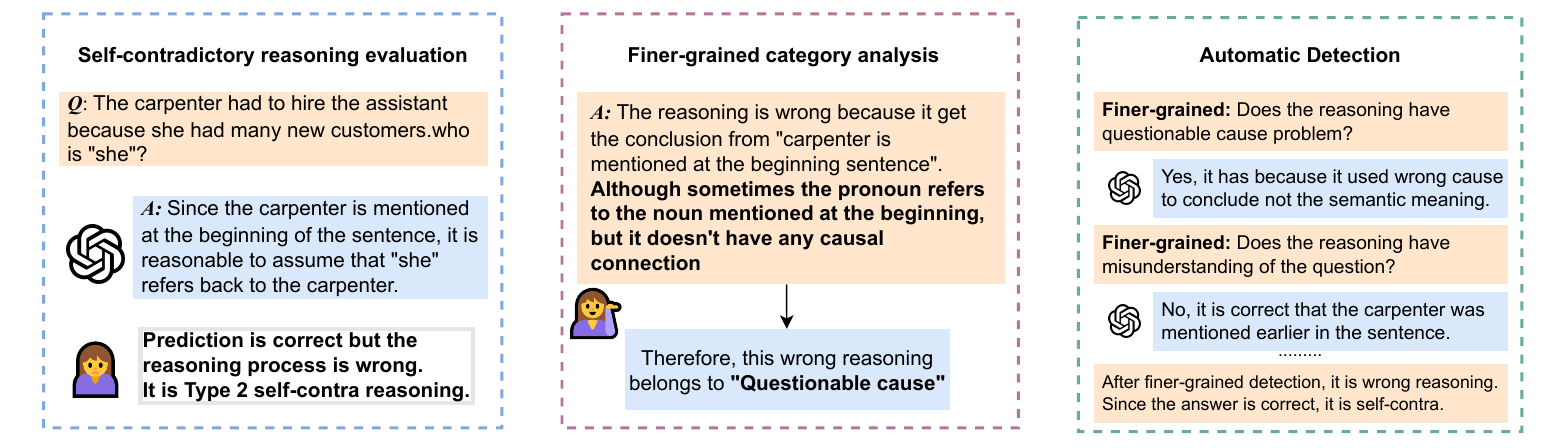}
   \caption{Three paradigms we study: human-annotated \sca reasoning evaluation, finer-grained category analysis, and finer-grained categories enhanced automatic detection of \sca. We first identify the type of \sca reasoning and analyze the detailed cause of the issues. Then we build automatic evaluation based on finer-grained category detection.}
   \vspace{-0.3cm}
    \label{fig:pipeline}
\end{figure*}


Many recent studies have explored the unfaithfulness in the reasoning ability of LLMs \cite{huang2023large,zheng2023does,ye2022unreliability,wiegreffe2020measuring,sanyal2024machines}. They demonstrated that models may fail to generate factual and consistent explanations.
 Although previous work has discussed inconsistency between reasoning and prediction, the specific mechanisms underlying how reasoning contributes to these inconsistencies remain unclear. Therefore, a thorough and comprehensive evaluation is crucial to dissect the logical fallacies inherent in the reasoning process.


In this work, we shift the paradigm of reasoning evaluation by \textsc{Self-Contra} reasoning in question answering tasks (Figure \ref{fig:intro}). 
We define \textsc{Self-Contra} reasoning as follows: correct reasoning leading to the wrong answer, wrong reasoning leading to the correct answer, or reasoning itself is self-contradictory. As shown in Figure \ref{fig:pipeline}, our study consists of 3 parts: (1) \textbf{\textsc{Self-Contra} Reasoning Evaluation:} we ask a model to generate an answer along with reasoning and evaluate if reasoning supports its prediction; (2) \textbf{Finer-grained Categories Analysis:} we create detailed categories within \textsc{Self-Contra} reasoning to specify the causes that contribute most to reasoning failures;  (3) \textbf{Automatic Detection of \sca reasoning:} we propose a method leveraging finer-grained categories to 
automatically evaluate \sca reasoning. 

We conduct \textsc{Self-Contra} reasoning evaluations on 3 datasets: WinoBias \cite{zhao2018gender}, WinoGrande \cite{sakaguchi2021winogrande}, WinoGender \cite{rudinger-EtAl:2018:N18} using 3 LLMs: GPT-3.5 \cite{gpt3},  Mistral 7B \cite{jiang2023mistral} and Claude3-opus \cite{claude3}. 
We find that \sca commonly exists in all 3 datasets and LLMs, e.g., 30\% in the WinoBias dataset using GPT-3.5. While high accuracies of reasoning tasks overall correlate with low \sca rates, models like GPT-3.5 deviate from this trend, underscoring the limitations of evaluating models based solely on task performance.
When correct reasoning yields wrong answers, the models usually conclude the answer before they generate a complete reasoning chain. Conversely, when wrong reasoning yields correct answers, models exhibit logical fallacies rather than engaging with the semantic context.  

\begin{table*}[h!]
\centering
\scalebox{0.60}{
\begin{tabular}{lll}
\toprule
    \textbf{Categories} & \textbf{Example} & \textbf{Comment}  \\\midrule
    \makecell[l]{
      Correct reasoning ->\\ wrong prediction (Type1)
    } & \makecell[l]{
    \textit{\colorbox{pink}{Question:}} The guard was a big fan of the writer because he was the best at \\writing romantic novels. who is ``he''? \\
    \textit{\colorbox{lightblue}{Reasoning}}: (1) The sentence states that the guard was a big fan of the writer. \\(2) A fan is someone who admires and enjoys the work of another person. \\(3) In this case, the guard admires and enjoys the writer's work, specifically his \\romantic novels.  \\
    \textit{\colorbox{lightblue}{Answer:}} ``He'' refers to the guard.}
    &  \makecell[l]{The reasoning is correct and indicates the writer\\ wrote the romantic novels but failed to make the \\correct prediction.}\\ 
    \midrule
        \makecell[l]{
    Wrong reasoning -> \\correct prediction (Type2)
    } & \makecell[l]{
    \textit{\colorbox{pink}{Question:}}  The analyst awkwardly ran into the attendant who kept raising\\ his voice with customers. Who does ``his'' refer to?\\
    \textit{\colorbox{lightblue}{Reasoning:}} (1) The pronoun ``his'' is used to indicate possession or association\\ \textcolor{red}{ with a male} person, (2) thus it is referring to the attendant who kept raising\\ his voice with customers.\\
    \textit{\colorbox{lightblue}{Answer:}} ``His'' refers to the attendant.}
    &  \makecell[l]{The reasoning is lack of logic, only using\\ male characteristics to predict the answer.
    }\\ 
    \midrule
    \makecell[l]{
    Reasoning itself \\ is self-contradictory (Type3)
    } & \makecell[l]{
    \textit{\colorbox{pink}{Question:}}  The cleaner got mad at the farmer because she brought lots\\ of dirt. Who is ``she''? \\
    \textit{\colorbox{lightblue}{Reasoning:}} (1) The cleaner got mad at the farmer because \textcolor{red}{the farmer brought}\\ lots of dirt. (2) Since the cleaner is the one who got mad, (3) it can be inferred \\that ``she'' refers to the cleaner.\\
    \textit{\colorbox{lightblue}{Answer:}} ``She'' refers to the cleaner.}
    &  \makecell[l]{The first sentence already stated that it is the \\farmer who brought lots of dirt but changed \\its answer to the cleaner in the end.
    }\\ 
\bottomrule
\end{tabular}
}
\caption{Examples of self-contradictory reasoning of each type from the WinoBias Dataset: we show the original question, generated results, and human comments on self-contradiction of the reasoning process. We mark reasoning steps for each reasoning.}
\vspace{-0.3cm}
\label{tab:examples}
\end{table*}

We then used GPT-4 in our study for automatic detection of \textsc{Self-Contra} reasoning since it has been demonstrated as a strong evaluator in the literature (\citet{naismith-etal-2023-automated,openai2023gpt4,hsu2023gpt4}).  
We compare finer-grained categories enhanced detection 
with other two baselines: (1) models predicting answers based on the concatenation of input question and model reasoning; (2) binary prediction using chain-of-thought prompting given demonstrations of \sca.
Our results show that detection based on finer-grained categories outperforms the other two baselines by 10-15\% in $F_1$ score, showing the efficacy of incorporating finer-grained analysis. However, GPT-4 performs notably worse than human detection, approximately 15\%  lower on average. For future research, we introduce \sca reasoning detection as a new task 
to assess the model's capability to identify problematic reasoning. This task is crucial, as an inability to identify logical fallacies hinders the generation of sound reasoning.

In summary, our key contributions are:

\begin{compactitem}
    \item We introduce the concept of \sca reasoning and provide the formal definition.
    \item We provide analysis on \sca reasoning from different granularity: we begin with a high-level assessment of disparity between prediction and reasoning, then progress to a finer-grained understanding of the causes of \sca reasoning.
    \item We introduce a new task: \sca reasoning detection task and our results underscore the continued challenge for most state-of-the-art models in this domain.
\end{compactitem}




\section{\textsc{Self-Contra} Reasoning}
\label{sec:self-contra}

We begin by defining \sca reasoning and then introduce the methods to probe such problematic reasoning in LLMs.
\subsection{Definition}
\label{sec:self-contra:definition}
In a self-rationalization setting where models generate reasoning with their output~\cite{marasovic2021few},
we can define self-contradictory reasoning using three categories:  \textbf{Type1}: a correct reasoning leading to a wrong prediction; \textbf{Type2}: a wrong reasoning leading to a correct prediction; \textbf{Type3}: there are contradictions in the reasoning itself. We consider reasoning as correct only when there is no wrong information or logical fallacy. Conversely, if any segment of reasoning is wrong, it will be deemed incorrect. Examples of each category are shown in Table~\ref{tab:examples}. \looseness=-1

In this paper, we define the reasoning generated by LLMs as a complete reasoning chain including premise, inference, and conclusion. 
Formally, let \(r\) be the reasoning, and \(a\) be the binary indication of the predicted answer being correct ($a=1$) or wrong ($a=0$). 
Note that one reasoning \(r\) could have $k (k\geq1)$ steps. We set $r_i$ to 1 to denote the $i$-th step is correct and 0 otherwise. Therefore, the formal definition of \textsc{Self-Contra} reasoning is:
\begin{equation*}
\small \textsc{Self-Contra} \coloneqq
\begin{cases}
\begin{aligned}
 \textsc{Type1} & \quad \text{if } \forall i, r_i = 1 \& a = 0 \\
 \textsc{Type2} & \quad \text{if } \exists i, r_i=0 \& a=1 \\
 \textsc{Type3} & \quad \text{if } \exists i\neq j,  r_i \text{ contradicts }r_j
\end{aligned}
\end{cases}
\end{equation*}
Besides, we denote a right reasoning leading to a right answer case as \textsc{RR} and a wrong reasoning leading to a wrong answer as \textsc{WW}.


\subsection{Dataset}
We use 3 datasets from different settings: WinoBias~\cite{zhao2018gender}, WinoGrande ~\cite{sakaguchi2021winogrande}, WinoGender \cite{rudinger-EtAl:2018:N18}. We choose these datasets as they evaluate different model capabilities: social bias detection and commonsense reasoning. 
We selected these datasets to differentiate from current reasoning work: first, the datasets consist of very short sentences or questions; second, they do not require any high-level knowledge and are very easy for humans to answer. Our goal is to focus on where models make reasoning mistakes on very simple tasks and to understand the reasons behind these errors.
Since studying self-contradictory requires rigorous annotation by the experts,
we first conduct experiments on a small set for each dataset (50 instances) and later expand experiments based on those sets. All the prompt templates we used and dataset details can be found in Appendix Sec.~\ref{sec:appendix:section3}. 

\subsection{Probing Reasoning in LLMs}
We consider different settings to understand to what extent LLMs can do the reasoning. For all the prompting methods, we use 3 LLMs: GPT-3.5-turbo \cite{gpt3}, Mistral 7B Instruct v0.2 \cite{jiang2023mistral}, and Claude3-opus \cite{claude3}. We set the temperature to 0 for all models (more details in Appendix Sec.~\ref{sec:appendix:section3}). 
\paragraph{Zero-  and Few-shot prompting}
We begin with zero-shot and few-shot prompting. We employ a dual approach, where we request the model to provide reasoning before delivering an answer (donated with `(R)'), and vice versa (denoted with `(A)'). In the few-shot prompting, we adopt  Chain-of-Thought prompting~\cite{wei2022chain}, which combines a manual curation of six instructional demonstrations including questions, human written reasonings, and answers.
Therefore, we conduct experiments using four prompt settings and three LLMs across three datasets, with each dataset comprising 50 samples. This results in a total of 1800 data points.\looseness=-1

\begin{table}
\centering
\scalebox{0.70}{
\begin{tabular}{lllrrr}
\toprule
\textbf{Model}&\textbf{Prompt}&\textbf{Metrics}&\textbf{WB}&\textbf{WG}&\textbf{WGr}\\
\midrule
\multirow{4}{*}{GPT-3.5}&\multirow{2}{*}{Zero}&Acc&0.56&0.86&0.74\\
&&SCR&0.28&0.60&0.30\\
&\multirow{2}{*}{Few}&Acc&0.72&0.96&0.78\\
&&SCR&0.32&0.26&0.18\\
\midrule
\multirow{4}{*}{Mistral 7B}&\multirow{2}{*}{Zero}&Acc&0.44&0.82&0.72\\
&&SCR&0.48&0.42&0.24\\
&\multirow{2}{*}{Few}&Acc&0.36&0.84&0.74\\
&&SCR&0.50&0.26&0.26\\
\midrule
\multirow{4}{*}{Claude 3}&\multirow{2}{*}{Zero}&Acc&0.78&0.86&0.94\\
&&SCR&0.10&0.12&0.06\\
&\multirow{2}{*}{Few}&Acc&0.86&1.00&1.00\\
&&SCR&0.10&0.02&0.04\\
\bottomrule
\end{tabular}
}
\caption{We report accuracies and SCR on the answer-first (A) setting. For the reason-first (R) setting, the results are in Appendix Sec.~\ref{sec:appendix:section3:result}. WB stands for WinoBias, WG for WinoGender, and WGr for WinoGrande. }
\vspace{-0.3cm}
\label{tab:answer_first_results}
\end{table}

\subsection{Results and Analysis}

We first report model accuracy and \textsc{Self-Contra} rate (SCR) for results where
   $\text{SCR} = \frac{\# \textsc{Self-Contra}}{\#\text{Total}}$.
We observe \sca commonly exists in LLM reasoning, especially in the zero-shot setting. Since reason-first (R) setting results share a similar pattern as the answer-first (A) setting, we report the reason-first (R) setting in Appendix Sec.~\ref{sec:appendix:section3:result}.
\paragraph{Which tasks and LLMs are prone to formulate \textsc{Self-Contra} reasoning?} 
 As shown in Table \ref{tab:answer_first_results}, all 3 datasets show \sca reasoning to some extent. WinoBias and WinoGender generally exhibit more \sca than WinoGrande, indicating that reasoning with social biases still remains a challenge to models while inherently easy for humans. This emphasizes the importance of robust reasoning to avoid reinforcing stereotypes in real-world interactions. Mistral 7B typically shows higher SCR than the other two models, while Claude 3 almost perfectly manages reasoning tasks with a very low SCR, particularly in the few-shot setting. Hence, larger-sized models appear to mitigate \sca issues effectively. However, even the state-of-the-art model cannot completely eliminate \sca rate, indicating that robust reasoning remains a challenge for LLMs.
\paragraph{Does accuracy correlate with SCR?} We compute the Pearson correlation coefficient \cite{sedgwick2012pearson} between accuracy and SCR across all datasets and LLMs. The correlation coefficient and p-value are -0.634 and 0.006 respectively, indicating a strong negative relationship; higher accuracy is associated with a lower SCR. However, GPT-3.5 and Mistral 7B do not share the similar pattern, where the p-values are 0.676 and 0.083, respectively. A similar pattern is also observed in the reason-first results (Appendix Sec.~\ref{sec:appendix:section3:result}). Therefore, contrary to existing literature, we argue that evaluating a model's reasoning ability should go beyond performance metrics like accuracy, which can overlook critical flaws in LLMs' reasoning.
\begin{figure*}
\centering
\begin{subfigure}[b]{0.47\textwidth}
  \centering
  \includegraphics[width=\textwidth]{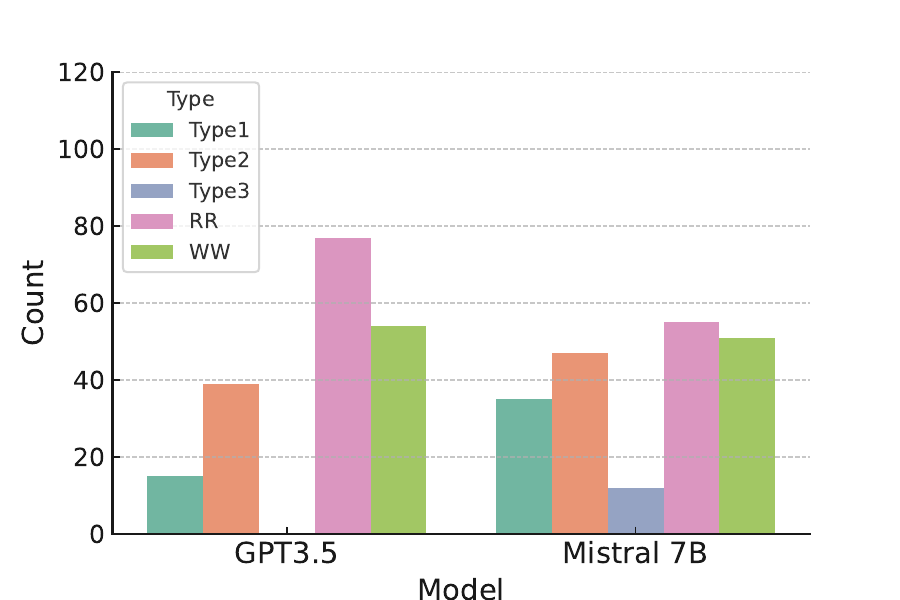}
  \caption{Results for WinoBias.}
  \label{fig:sub1}
\end{subfigure}
\hfill
\begin{subfigure}[b]{0.47\textwidth}
  \centering
  \includegraphics[width=\textwidth]{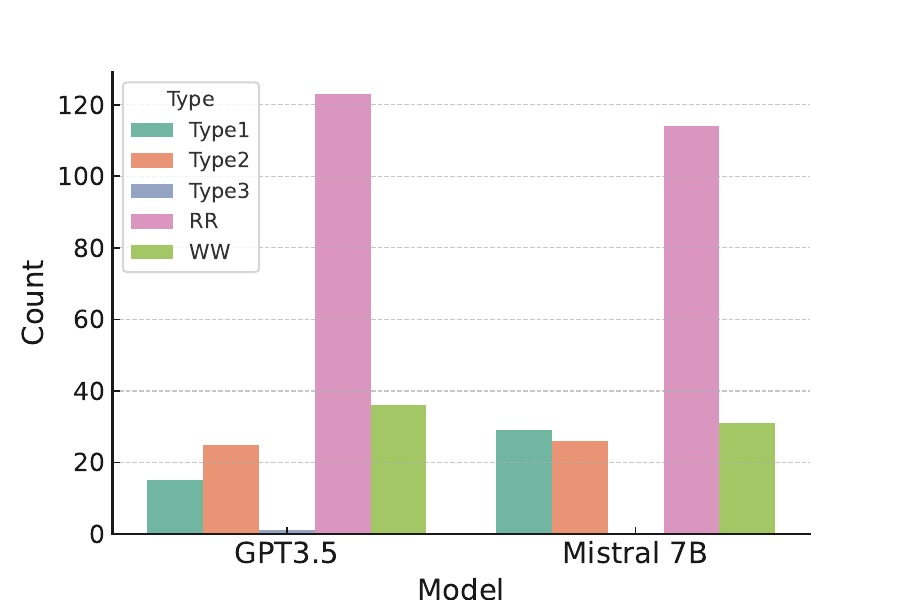}
  \caption{Results for WinoGrande. }
  \label{fig:sub2}
\end{subfigure}
\caption{Frequency of types in WinoBias and WinoGrande datasets. The result of WinoGender dataset is shown in Appendix Sec.~\ref{sec:appendix:section3:result}. We combine zero-shot and few-shot results. \textbf{Takeaway:} Type 2 reasoning accounts for a large portion of \sca which could hurt users' confidence in LLMs as wrong reasoning yields correct answers.
}
\vspace{-0.3cm}
\label{fig:type}
\end{figure*}

 \paragraph{Which are the most common reasoning?}
 As shown in Figure \ref{fig:type}, we found that in the Mistral 7B setting, the lack of significant difference between Type 2 and WW indicates that incorrect reasoning might randomly lead to either correct or incorrect answers, while GPT-3.5 demonstrates a bit better performance in distinguishing correct from incorrect reasoning. Type 1 errors are generally rarer than RR cases, indicating that correct reasoning tends to lead to the correct answer. This pattern emphasizes the challenge of robust reasoning, particularly for Type 2 errors: even when the answers are correct, faulty reasoning can undermine user confidence in the model. 

 Overall, our results demonstrate that \sca remains a significant problem for LLMs and that relying solely on performance-based metrics may obscure underlying issues in their reasoning capabilities.

\section{Finer-grained Categories of \textsc{Self-Contra}}
\label{sec:finer}

During the evaluation, we observed that the model makes the same type of mistakes multiple times over the dataset, such as logical fallacy, wrong knowledge, only using shortcuts to reason, details missing, etc. To understand why these models make self-contradictory reasoning, we delve into those problematic reasoning in detail and categorize them into finer-grained types. Below, we provide the definitions for the categories and put all the examples in Table~\ref{tab:finer_examples_main}. 
\begin{table*}[ht]
\centering
\scalebox{0.69}{
\begin{tabular}{lll}
\toprule
    \textbf{Categories} & \textbf{Example} & \textbf{Comment}  \\\midrule
    \makecell[l]{
      Evidence missing 
    } & \makecell[l]{
    \textit{\colorbox{pink}{Question:}} The carpenter had to hire the assistant because he was a friend \\of a friend. Who is ``he''? \\ 
    \textit{\colorbox{lightblue}{Reasoning:}} In the given sentence, ``he'' refers to the carpenter because the \\sentence states that \textcolor{red}{the carpenter had to hire the assistant.}\\
    \textit{\colorbox{lightblue}{Answer:}} The carpenter}
    &  \makecell[l]{Reasoning part only includes the first clause. \\Reasoning itself does not carry wrong\\ information but fails to reason based on the \\whole sentence.}\\ 
    \midrule
        \makecell[l]{
    Incomplete reasoning 
    } & \makecell[l]{
    \textit{\colorbox{pink}{Question:}} The carpenter had to hire the assistant because he was a friend \\of a friend. who is ``he''? \\
    \textit{\colorbox{lightblue}{Reasoning:}} The sentence states that the carpenter had to hire the assistant \\because he was a friend of a friend. This means that the carpenter was the \\one who hired the assistant based on their mutual connection.\\ 
    \textit{\colorbox{lightblue}{Answer:}} The carpenter}
    &  \makecell[l]{The reasoning correctly interprets the \\question. However,it missed final reasoning \\step by  linking ``assistant'' to ``he.''}\\ 
    \midrule
    \makecell[l]{
    Questionable cause 
    } & \makecell[l]{
    \textit{\colorbox{pink}{Question:}}  The farmer demanded more bread from the baker because she \\was hungry. Who is ``she''? \\
    \textit{\colorbox{lightblue}{Reasoning:}} \textcolor{red}{Since the pronoun ``she'' is used,} it can be inferred that the \\person referred to as ``she'' is the farmer.\\
    \textit{\colorbox{lightblue}{Answer:}} ``She'' refers to the farmer}
    &  \makecell[l]{Reasoning used the wrong cause (``She'' is \\used) to conclude ``she'' refers to the farmer.}\\ 
    \midrule
    \makecell[l]{
    Begging the question
    } & \makecell[l]{
    \textit{\colorbox{pink}{Question:}}  The analyst walked to the cashier because she needed to make \\a deposit. Who is ``she''?\\
    \textit{\colorbox{lightblue}{Reasoning:}} \textcolor{red}{Since the analyst is the one who needed to make a deposit}, \\``she'' refers to the analyst.\\
    \textit{\colorbox{lightblue}{Answer:}} Analyst\\}
    &  \makecell[l]{The reasoning already assumes the truth of \\the answer and uses conclusion to reason. }\\ 
    \midrule
    \makecell[l]{
    Wrong context\\ knowledge
    } & \makecell[l]{
    \textit{\colorbox{pink}{Question:}} The guard was a big fan of the writer because she loved \\romantic novels. Who is ``she''? \\
    \textit{\colorbox{lightblue}{Reasoning:}} \textcolor{red}{Since the guard is the only person mentioned in the sentence},\\ it can be inferred that ``she'' refers to the guard.\\
    \textit{\colorbox{lightblue}{Answer:}} The guard}
    &  \makecell[l]{The guard is not the only person mentioned \\in the sentence. The model captures the \\wrong context information.}\\

\bottomrule
\end{tabular}
}
\caption{\textbf{Examples of finer-grained categories from the WinoBias Dataset}: We show the original question, generated results, and human comments. }
\label{tab:finer_examples_main}
\end{table*}
\subsection{Correct Reasoning Categories} 
 We define good reasoning as complete, accurate, and logically sound, while bad reasoning includes wrong information, logical fallacies, or incomplete reasoning. We note that, for both Type1 and RR, all reasoning segments are considered correct under our definition. Moreover, \textit{correct reasoning does not equal good reasoning}. Imperfectly correct reasoning can still have the following issues:
\begin{compactitem}
    \item \textbf{Evidence missing (EM)} The model only generates reasoning based on partial evidence. 
    \item \textbf{Incomplete reasoning (IR)} The model captures all evidence and follows some sound reasoning but fails to link its prediction to its reasoning.
\end{compactitem}
\begin{table}
\centering
\scalebox{0.8}{
\begin{tabular}{cccc}
\toprule
Category&Self-Contra&Type & Finer-grained\\
\midrule
Krippendorff's $\alpha$ & 0.89&0.93&0.89\\
\bottomrule
\end{tabular}
}
\caption{Annotator agreement on Anti-dev dataset.}
\vspace{-0.5cm}
\label{tab:agreement}
\end{table}

\begin{table*}[h!]
\centering
\scalebox{0.7}{
\begin{tabular}{llcccccccc}
\toprule
\multirow{2}{*}{ \textbf{Models}}&\multirow{2}{*}{  \textbf{Dataset}}&\multicolumn{2}{c}{ \textbf{ \makecell[c]{Correct Reasoning \\(Type 1)}}}&\multicolumn{3}{c}{ \textbf{\makecell[c]{ Wrong Reasoning \\(Type 2)} }}&\multirow{2}{*}{ \textbf{\makecell[c]{Self-Contra \\(Type 3)}}}&\multirow{2}{*}{  \textbf{Total Count}}\\
\cmidrule{3-4} \cmidrule(lr){5-7}
& &   \makecell[c]{EM}&  \makecell[c]{IR}&  \makecell[c]{QC}&  \makecell[c]{BQ}&  \makecell[c]{WCK}&  \\
 \midrule
 \multirow{3}{*}{ \textbf{GPT-3.5}}
 &  \textbf{WB Anti (test)}&  0.08   &   0.168  &  0.027  &   0.671   &   0.024 &  0.008 &  129 \\

 &  \textbf{WB Pro (test)} &   0.000   &  0.046  &   0.144  &   0.800  &  0.026 & 0.000 & 196\\
&  \textbf{WinoGrande} &   0.000  &  0.248  &  0.016 &  0.624   &  0.016 &  0.080 &  90\\

 \midrule
 \multirow{3}{*}{ \textbf{Mistral 7B}}
 &  \textbf{WB Anti (test)}&  0.030  &   0.454 &  0.018 &   0.282  &   0.026  &   0.209 &  163\\

 &  \textbf{WB Pro (test)} &   0.000  &  0.094  &   0.112 &   0.720  & 0.017 &  0.008 &  113\\&
   \textbf{WinoGrande} &   0.000  &   0.248 &  0.016 &   0.624  &   0.016&  0.080 &  117\\
\bottomrule
\end{tabular}
}

\caption{Ratio of finer-grained categories. We also include Type3 results as \sca cases consist of all 3 types. The last column is the total count of \sca cases for each dataset. For example, there are 129 \sca cases in GPT-3.5 WB Anti (test) results over 353 data points in total. } 
\label{tab:finer-grained results}
\end{table*}

\subsection{Wrong Reasoning Categories}
We define \textsc{Type2} reasoning as wrong reasoning leading to a correct answer. In this case, the reasoning does not follow a logical reasoning path but uses shortcuts or syntactical rules. The 3 common categories of wrong reasoning are as follows:
\begin{compactitem}

    \item \textbf{Questionable Cause (QC)} Also known as a causal fallacy, questionable cause \cite{petric2020logical} is a category of informal fallacy in which a cause is incorrectly identified. 
  \item \textbf{Begging the Question (BQ)} The fallacy of begging the question \cite{barker1976fallacy} occurs when an argument's premises assume the truth of the conclusion, instead of supporting it. 
   \item \textbf{Wrong context knowledge (WCK)} The model interprets the input information wrongly. \looseness=-1
\end{compactitem}
    

    
    \subsection{Results}
    In Section \ref{sec:self-contra}, Claude 3 exhibits less \sca than other models so we only focus on GPT-3.5 and Mistral 7B in this section. Additionally, we report results on WinoBias and WinoGrande since WinoGender is similar to WinoBias. Among the 4 prompt settings, the Few-shot (A) setting demonstrates the best performance with a high accuracy and a low SCR. 
    WinoBias includes ``pro-stereotype'' where examples follow US social stereotypes, and ``anti-stereotype'' where examples are against the stereotypes. We conduct analysis on both ``pro'' and ``anti'' settings from the WinoBias test set. Thus, our subsequent analysis primarily concentrates on the WinoBias and WinoGrande datasets using GPT3.5 and Mistral 7B with a Few-shot (A) setting. The WinoBias and WinoGrande datasets include 353 and 357 data points, respectively. Overall, the finer-grained analysis includes 2126 data points annotated by human experts (more details of annotations are in Appendix Sec.~\ref{sec:finer}). Since annotation requires a strong understanding of what constitutes \sca reasoning, we opted to self-annotate the data between two of the authors. The annotators first annotated the same 200 samples from the dataset with 0.89 inter-annotator agreement as shown in Table \ref{tab:agreement}. Following \citet{ye2022unreliability}, each instance was annotated by one author only. 
    
    We report the results of all categories in Table~\ref{tab:finer-grained results}. For each category, we report the ratio of each category over \sca cases. In Type 1 cases, the majority stem from incomplete reasoning, with only a few due to evidence missing. This can be largely attributed to few-shot prompting, which trains the model to utilize all available evidence in its reasoning. We extend our experiments on the WinoBias dataset to include various prompting methods, with detailed results presented in Appendix Sec.~\ref{sec:app:finer}. Notably, evidence missing is more prevalent in the zero-shot setting, where models frequently generate brief reasoning and overlook evidence. In Type 2 cases, the most common issue is begging the question, while errors such as questionable cause and wrong context knowledge are relatively infrequent. This pattern also relates to few-shot prompting as in zero-shot settings, models tend to use shortcuts and follow the wrong pattern consistently to reason, but few-shot demonstrations encourage the model to reason using semantics. Overall, issues like evidence missing, questionable cause, and wrong context knowledge are more readily mitigated by models after learning through demonstrations, as these patterns are relatively straightforward — either overlooking/misunderstanding evidence or relying solely on the syntax for reasoning. However, overcoming the fallacy of begging the question remains challenging for models. Incomplete reasoning is often the result of models losing focus during the reasoning process. As shown in Table \ref{tab:finer_examples_main}, if a sentence involves two characters, the model may focus its reasoning on the first character and then prematurely conclude with an answer about this character, failing to establish a connection to the second character.


\section{Automatic detection}

Previous evaluations rely solely on human annotation. Exploring whether models can detect \sca reasoning could significantly reduce human workload and benefit the community. In this section, we explore the capability of LLMs in detecting \sca. 
\begin{table*}
\centering
\scalebox{0.7}{
\begin{tabular}{llccccccc}
\toprule
\textbf{Model}&\textbf{Dataset} & \multicolumn{3}{c}{\textbf{\sca detection}}&\multicolumn{4}{c}{\textbf{Finer-grained categories detection}}\\
\cmidrule{3-5} \cmidrule (lr){6-9}
&&\textbf{I+R$\rightarrow$O}&\textbf{Binary}&\textbf{FGE}&\textbf{QC}&\textbf{BQ}&\textbf{WCK}&\textbf{SC}\\
\midrule
\multirow{3}{*}{\textbf{GPT-3.5}}&\textbf{WB Anti (test)}&0.208 &0.253 &0.522 &0.279&0.300&0.000&0.031\\
&\textbf{WB Pro (test)}&0.065 &0.233 &0.628&0.201&0.431&0.000&-\\
&\textbf{WinoGrande}& 0.289&0.360&0.504&0.000&0.534&0.000&0.116\\
\midrule
\multirow{3}{*}{\textbf{Mistral 7B}}&\textbf{WB Anti (test)}&0.584 & 0.595&0.543&0.060&0.339&0.000&0.284\\
&\textbf{WB Pro (test)}&0.154 &0.215 &0.454&0.059&0.328&0.000&0.273\\
&\textbf{WinoGrande}&0.328 & 0.395&0.484&0.000&0.446&0.042&0.182\\
\bottomrule
\end{tabular}
}
\caption{Automatic detection of \sca and finer-grained categories. We report $F_1$ scores in the table and ``-'' means there is no such category in the dataset. We use GPT-4 to evaluate reasoning generated by GPT-3.5 and Mistral 7B.}
\label{tab:automatic_detect}
\end{table*}
\subsection{Methods}
\paragraph{I+R$\rightarrow$O} 
An intuitive method to evaluate the faithfulness of reasoning involves incorporating the reasoning directly into the prompt \cite{wiegreffe2020measuring}. We prompt models to respond to a question based on the reasoning they themselves generate. Since this reasoning typically includes the conclusions, we omit these from the prompt to focus solely on the reasoning process. If the new prediction changes from the original one after the model is given the reasoning, we consider it as \sca reasoning.
\paragraph{Binary detection} We directly prompt  GPT-4-turbo model to produce a binary prediction about whether the reasoning is \textsc{Self-Contra} using six demonstrations, with three non-\sca cases and three \sca cases. 

\paragraph{Finer-Grained categories Enhanced (FGE) \sca detection} We ask the GPT-4 model to predict the finer-grained category for the whole reasoning path given the definition of each wrong reasoning finer-grained category. We then calculate the type and \sca reasoning based on these finer-grained category predictions. The result is calculated based on the definition in Section \ref{sec:finer} as follows:
\begin{equation*}
\small \textsc{Self-Contra} \coloneqq
\begin{cases}
\begin{aligned}
 \textsc{Type1} & \quad \text{if } \forall i, w_i \neq 1 \& a = 0 \\
 \textsc{Type2} & \quad \text{if } \exists i, w_i=1 \& a=1 \\
\end{aligned}
\end{cases}
\end{equation*}
where $w$ is an indicator representing if the model detects certain wrong reasoning categories ($w$ = 1) or the model does not detect them ($w$ = 0). $a$ denotes if the prediction is correct ($a$ = 1) or wrong ($a$ = 0) and $i$ denotes wrong reasoning finer-grained category id. Note that, in Type1 all the reasoning segments are correct. As long as GPT-4 does not predict any wrong category (e.g. questionable cause) in the reasoning, we consider the reasoning as correct. Given that Type3 is not part of the finer-grained category, our approach directly asks GPT-4 if the reasoning itself is self-contradictory, distinct from the binary setting which asks for all the \sca types.
If the model responds yes, we classify the reasoning as \sca.
We implement an ensembled predictor based on finer-grained category prediction where GPT-4 is prompted to give a binary prediction for each category and we ensemble the results according to the above formula. We provide the definition and 6 demonstrations for each category. All the prompt templates we used in this section are shown in Appendix Sec.~\ref{sec:prompts_automatic} and all prompts are fine-tuned multiple times till we get a desirable result.


\paragraph{Human detection} Four computer science students volunteer to annotate under the same setting
as FGE. Each annotator is given a definition and 6 demonstrations of each finer-grained category and annotates SCR and finer-grained categories of 150 samples from the WinoBias anti-test set.  This setting is designed to compare with the model, aiming to determine whether the task is inherently challenging or if the model's capability is limited.


\subsection{Results}
\paragraph{FGE detection generally outperforms the other two methods but still is worse than human performance.}As demonstrated in Table \ref{tab:automatic_detect}, the FGE detection
surpasses the other two methods in performance, with binary detection proving more effective than the I+R$\rightarrow$O methods. This indicates that using finer-grained categories for automatic evaluation significantly enhances the model's ability to identify \sca reasoning. However, binary detection, which utilizes only six demonstrations for distinguishing between \sca and non \sca cases, may not provide sufficient learning material for models. The I+R$\rightarrow$O method struggles for two main reasons: first, it fails to identify logical fallacies such as begging the question—where the reasoning, although wrong, presumes the conclusion within the premise, leading to the same answer when the model is queried; second, it often encounters cases with incomplete reasoning or evidence missing, which can result in either correct or incorrect answers (RR and Type1 respectively). Thus, instances of incomplete reasoning in RR cases might also yield incorrect conclusions, highlighting a lack of robustness in reasoning. However, I+R$\rightarrow$O performs efficiently for Mistral 7B in the WinoBias Anti test set. The main reason behind this is that Type 1 errors, which constitute two-thirds of the \sca cases, are mostly caused by incomplete reasoning. In such scenarios, the I+R$\rightarrow$O model can perform well. Despite the FGE's superior performance, a state-of-the-art model like GPT-4 struggles with detecting \sca, achieving an $F_1$ score of approximately 0.5. In contrast, human annotators achieved an average $F_1$ score of 0.667 on the anti-test set.


\paragraph{GPT-4 detects BQ more effectively than others. }We further investigate the finer-grained detection performance of the FGE setting across the complete datasets, which include both \sca and non-\sca cases. As illustrated in Table \ref{tab:automatic_detect}, the model more effectively detects instances of begging the question compared to other categories, but it performs poorly on categories such as wrong context knowledge. The predominance of the begging the question category in the data allows for more effective fine-tuning of the prompt, thereby enhancing performance. However, for less frequent categories like wrong context knowledge and questionable cause—sometimes represented by only 2-3 cases in the dataset—the model struggles to detect these errors. Additionally, the tendency of models to predict ``0'' for wrong context knowledge underscores their limited ability to identify hallucinations or incorrect information.





\section{Related Work}

\paragraph{Inconsistency and unfaithfulness of LLM in reasoning}
There has been extensive current work on the hallucination and faithfulness of LLM reasoning. \citet{turpin2023language} demonstrates that CoT explanations can be plausible yet systematically unfaithful. \citet{mundler2023self} shows that LLM can generate two self-contradictory claims toward the same entity. LLMs are also fragile when faced with simple challenges, often changing their answers quickly \cite{laban2024surechallengingllmsleads}.
Many works have stated that LLMs' rationale does not completely support labels \cite{wiegreffe2020measuring,ye2022unreliability}. \citet{wang2022towards} studied how much valid reasoning matters and found that the inclusion of invalid reasoning did not significantly impact the accuracy of predictions.
Prior works proposed different techniques to improve reasoning and faithfulness in LLMs. \citet{ross2022does} trained model with human-written rationales to improve the robustness. \citet{lyu2023faithful} employed an LLM to translate a query into a chain of reasoning that can be executed deterministically. \citet{wang2022pinto} used counterfactual regularization to learn faithful reasoning over rationales. 
\citet{ramnath2023tailoring} used multi-reward to improve the rationale's plausibility. Moreover, self-consistency \cite{wang2022self}, chain-of-verification \cite{dhuliawala2023chain}, self-evaluation \cite{xie2023self}, multi-agent debate \cite{chan2023chateval}, chain-of-questions \cite{zhu2023chain}, and round-table conference reasoning \cite{chen2023reconcile} were proposed to improve the task performance by adding multiple reasoning steps. 
\paragraph{Self-Contradiction in LLMs} 

Previous literature has explored different kinds of self-contradictions \cite{hsu2021wikicontradiction,de-marneffe-etal-2008-finding,mündler2024selfcontradictory}, they mostly focus on the contradiction of factual knowledge between the contexts. \citet{ross2022does} measures the robustness of LLM reasoning against spurious correlations. \citet{zheng2023does} investigate the shortcomings of ChatGPT in truthful LLM reasoning. In contrast, our main focus is to examine the internal consistency between reasoning and predictions, particularly in cases where reasoning exhibits self-contradiction. 

\paragraph{Fine-grained reasoning evaluation}

\citet{golovneva2023roscoesuitemetricsscoring} proposed a suite of evaluation metrics for step-by-step reasoning. However, these metrics do not effectively assess the causality within the reasoning process or the relationship with the predicted answer. Similarly, \citet{jacovi2024chainofthoughtstrongweakestlink} introduced a new dataset and benchmark for evaluating chain-of-thought reasoning. Additionally, \citet{hao2024llmreasonersnewevaluation} designed different evaluation criteria tailored to various reasoning tasks. Our work concentrates on identifying specific reasoning fallacies within simple reasoning tasks, where models make clear mistakes, and we incorporate human annotations to capture these errors.

\section{Conclusion}
Our study focuses on \textsc{Self-Contra} reasoning in LLMs for question-answering tasks. We conduct \sca reasoning evaluation across 3 datasets and 3 LLMs. Next, we employ the WinoBias and WinoGrande datasets for an in-depth analysis. We analyze specific errors, such as models generating incomplete reasoning or using shortcuts, contributing to \sca reasoning. We also find that LLM is still not capable of detecting \sca reasoning, with a lower performance compared to humans. This work represents the first comprehensive study of \sca reasoning, offering a nuanced evaluation and a new task--automatic detection of \sca.

\section{Limitations}
Despite attempts at automatic evaluation, the performance is suboptimal, indicating a deficiency in the model's understanding of \sca reasoning. Future work should focus on enhancing the model's detection capabilities for \sca reasoning.
While we perform \sca evaluation across three datasets, our in-depth analysis is exclusively conducted on WinoBias and WinoGrande. Future efforts can extend this analysis to additional logical reasoning datasets to uncover further instances of reasoning errors.
Although finer-grained categories in this paper cover 95\% of reasoning contradiction errors, there are still some other logical fallacies we do not include.
\section{Ethics statement}
In order to build trustworthy models, we need to understand model behaviors better. Particularly, reasoning has serious potential to mislead people as LLMs become more and more fluent believable, but their reasoning is not necessarily factual or faithful. As a first step towards building trustworthy system for reasoning, our evaluative framework provides a tool for categorizing faulty reasoning that seek better behavioral understanding for transparency. We also experimented on a bias-conscious dataset, WinoBias, to test model's reasoning.
\section{Acknowledgement}

We would like to thank our collaborator, Kai-Wei Chang, for his constructive feedback on this work. The data annotation and human evaluation were conducted by undergraduate volunteers from USC. We especially appreciate the contributions of Misha Fu, Minhao Li, and Haosheng Gan for their assistance with the annotation and evaluation process.


\bibliography{anthology,custom}

\begin{thebibliography}{45}
\expandafter\ifx\csname natexlab\endcsname\relax\def\natexlab#1{#1}\fi

\bibitem[{Anthropic(2024)}]{claude3}
Anthropic. 2024.
\newblock Introducing the next generation of claude.
\newblock Technical report.

\bibitem[{Barker(1976)}]{barker1976fallacy}
John~A Barker. 1976.
\newblock The fallacy of begging the question.
\newblock \emph{Dialogue: Canadian Philosophical Review/Revue canadienne de philosophie}, 15(2):241--255.

\bibitem[{Bauer et~al.(2018)Bauer, Wang, and Bansal}]{bauer2018commonsense}
Lisa Bauer, Yicheng Wang, and Mohit Bansal. 2018.
\newblock Commonsense for generative multi-hop question answering tasks.
\newblock \emph{arXiv preprint arXiv:1809.06309}.

\bibitem[{Chan et~al.(2023)Chan, Chen, Su, Yu, Xue, Zhang, Fu, and Liu}]{chan2023chateval}
Chi-Min Chan, Weize Chen, Yusheng Su, Jianxuan Yu, Wei Xue, Shanghang Zhang, Jie Fu, and Zhiyuan Liu. 2023.
\newblock Chateval: Towards better llm-based evaluators through multi-agent debate.
\newblock \emph{arXiv preprint arXiv:2308.07201}.

\bibitem[{Chen et~al.(2023)Chen, Saha, and Bansal}]{chen2023reconcile}
Justin Chih-Yao Chen, Swarnadeep Saha, and Mohit Bansal. 2023.
\newblock Reconcile: Round-table conference improves reasoning via consensus among diverse llms.
\newblock \emph{arXiv preprint arXiv:2309.13007}.

\bibitem[{Chung et~al.(2022)Chung, Hou, Longpre, Zoph, Tay, Fedus, Li, Wang, Dehghani, Brahma et~al.}]{chung2022scaling}
Hyung~Won Chung, Le~Hou, Shayne Longpre, Barret Zoph, Yi~Tay, William Fedus, Yunxuan Li, Xuezhi Wang, Mostafa Dehghani, Siddhartha Brahma, et~al. 2022.
\newblock Scaling instruction-finetuned language models.
\newblock \emph{arXiv preprint arXiv:2210.11416}.

\bibitem[{de~Marneffe et~al.(2008)de~Marneffe, Rafferty, and Manning}]{de-marneffe-etal-2008-finding}
Marie-Catherine de~Marneffe, Anna~N. Rafferty, and Christopher~D. Manning. 2008.
\newblock \href {https://aclanthology.org/P08-1118} {Finding contradictions in text}.
\newblock In \emph{Proceedings of ACL-08: HLT}, pages 1039--1047, Columbus, Ohio. Association for Computational Linguistics.

\bibitem[{Dhuliawala et~al.(2023)Dhuliawala, Komeili, Xu, Raileanu, Li, Celikyilmaz, and Weston}]{dhuliawala2023chain}
Shehzaad Dhuliawala, Mojtaba Komeili, Jing Xu, Roberta Raileanu, Xian Li, Asli Celikyilmaz, and Jason Weston. 2023.
\newblock Chain-of-verification reduces hallucination in large language models.
\newblock \emph{arXiv preprint arXiv:2309.11495}.

\bibitem[{Golovneva et~al.(2023)Golovneva, Chen, Poff, Corredor, Zettlemoyer, Fazel-Zarandi, and Celikyilmaz}]{golovneva2023roscoesuitemetricsscoring}
Olga Golovneva, Moya Chen, Spencer Poff, Martin Corredor, Luke Zettlemoyer, Maryam Fazel-Zarandi, and Asli Celikyilmaz. 2023.
\newblock \href {http://arxiv.org/abs/2212.07919} {Roscoe: A suite of metrics for scoring step-by-step reasoning}.

\bibitem[{Hao et~al.(2024)Hao, Gu, Luo, Liu, Shao, Wang, Xie, Ma, Samavedhi, Gao, Wang, and Hu}]{hao2024llmreasonersnewevaluation}
Shibo Hao, Yi~Gu, Haotian Luo, Tianyang Liu, Xiyan Shao, Xinyuan Wang, Shuhua Xie, Haodi Ma, Adithya Samavedhi, Qiyue Gao, Zhen Wang, and Zhiting Hu. 2024.
\newblock \href {http://arxiv.org/abs/2404.05221} {Llm reasoners: New evaluation, library, and analysis of step-by-step reasoning with large language models}.

\bibitem[{Hsu et~al.(2021)Hsu, Li, Saez-Trumper, and Hsu}]{hsu2021wikicontradiction}
Cheng Hsu, Cheng-Te Li, Diego Saez-Trumper, and Yi-Zhan Hsu. 2021.
\newblock \href {http://arxiv.org/abs/2111.08543} {Wikicontradiction: Detecting self-contradiction articles on wikipedia}.

\bibitem[{Hsu et~al.(2023)Hsu, Huang, Rossi, Kim, Giles, and Huang}]{hsu2023gpt4}
Ting-Yao Hsu, Chieh-Yang Huang, Ryan Rossi, Sungchul Kim, C.~Lee Giles, and Ting-Hao~K. Huang. 2023.
\newblock \href {http://arxiv.org/abs/2310.15405} {Gpt-4 as an effective zero-shot evaluator for scientific figure captions}.

\bibitem[{Huang et~al.(2023)Huang, Chen, Mishra, Zheng, Yu, Song, and Zhou}]{huang2023large}
Jie Huang, Xinyun Chen, Swaroop Mishra, Huaixiu~Steven Zheng, Adams~Wei Yu, Xinying Song, and Denny Zhou. 2023.
\newblock Large language models cannot self-correct reasoning yet.
\newblock \emph{arXiv preprint arXiv:2310.01798}.

\bibitem[{Jacovi et~al.(2024)Jacovi, Bitton, Bohnet, Herzig, Honovich, Tseng, Collins, Aharoni, and Geva}]{jacovi2024chainofthoughtstrongweakestlink}
Alon Jacovi, Yonatan Bitton, Bernd Bohnet, Jonathan Herzig, Or~Honovich, Michael Tseng, Michael Collins, Roee Aharoni, and Mor Geva. 2024.
\newblock \href {http://arxiv.org/abs/2402.00559} {A chain-of-thought is as strong as its weakest link: A benchmark for verifiers of reasoning chains}.

\bibitem[{Jiang et~al.(2023)Jiang, Sablayrolles, Mensch, Bamford, Chaplot, de~las Casas, Bressand, Lengyel, Lample, Saulnier, Lavaud, Lachaux, Stock, Scao, Lavril, Wang, Lacroix, and Sayed}]{jiang2023mistral}
Albert~Q. Jiang, Alexandre Sablayrolles, Arthur Mensch, Chris Bamford, Devendra~Singh Chaplot, Diego de~las Casas, Florian Bressand, Gianna Lengyel, Guillaume Lample, Lucile Saulnier, Lélio~Renard Lavaud, Marie-Anne Lachaux, Pierre Stock, Teven~Le Scao, Thibaut Lavril, Thomas Wang, Timothée Lacroix, and William~El Sayed. 2023.
\newblock \href {http://arxiv.org/abs/2310.06825} {Mistral 7b}.

\bibitem[{Kojima et~al.(2022)Kojima, Gu, Reid, Matsuo, and Iwasawa}]{kojima2022large}
Takeshi Kojima, Shixiang~Shane Gu, Machel Reid, Yutaka Matsuo, and Yusuke Iwasawa. 2022.
\newblock Large language models are zero-shot reasoners.
\newblock \emph{Advances in neural information processing systems}, 35:22199--22213.

\bibitem[{Laban et~al.(2024)Laban, Murakhovs'ka, Xiong, and Wu}]{laban2024surechallengingllmsleads}
Philippe Laban, Lidiya Murakhovs'ka, Caiming Xiong, and Chien-Sheng Wu. 2024.
\newblock \href {http://arxiv.org/abs/2311.08596} {Are you sure? challenging llms leads to performance drops in the flipflop experiment}.

\bibitem[{Liu et~al.(2023)Liu, Yao, Ton, Zhang, Cheng, Klochkov, Taufiq, and Li}]{liu2023trustworthy}
Yang Liu, Yuanshun Yao, Jean-Francois Ton, Xiaoying Zhang, Ruocheng Guo~Hao Cheng, Yegor Klochkov, Muhammad~Faaiz Taufiq, and Hang Li. 2023.
\newblock Trustworthy llms: a survey and guideline for evaluating large language models' alignment.
\newblock \emph{arXiv preprint arXiv:2308.05374}.

\bibitem[{Lyu et~al.(2023)Lyu, Havaldar, Stein, Zhang, Rao, Wong, Apidianaki, and Callison-Burch}]{lyu2023faithful}
Qing Lyu, Shreya Havaldar, Adam Stein, Li~Zhang, Delip Rao, Eric Wong, Marianna Apidianaki, and Chris Callison-Burch. 2023.
\newblock Faithful chain-of-thought reasoning.
\newblock \emph{arXiv preprint arXiv:2301.13379}.

\bibitem[{Marasovi{\'c} et~al.(2021)Marasovi{\'c}, Beltagy, Downey, and Peters}]{marasovic2021few}
Ana Marasovi{\'c}, Iz~Beltagy, Doug Downey, and Matthew~E Peters. 2021.
\newblock Few-shot self-rationalization with natural language prompts.
\newblock \emph{arXiv preprint arXiv:2111.08284}.

\bibitem[{McCoy et~al.(2019)McCoy, Pavlick, and Linzen}]{mccoy2019right}
Tom McCoy, Ellie Pavlick, and Tal Linzen. 2019.
\newblock Right for the wrong reasons: Diagnosing syntactic heuristics in natural language inference.
\newblock In \emph{Proceedings of the 57th Annual Meeting of the Association for Computational Linguistics}, pages 3428--3448.

\bibitem[{M{\"u}ndler et~al.(2023)M{\"u}ndler, He, Jenko, and Vechev}]{mundler2023self}
Niels M{\"u}ndler, Jingxuan He, Slobodan Jenko, and Martin Vechev. 2023.
\newblock Self-contradictory hallucinations of large language models: Evaluation, detection and mitigation.
\newblock \emph{arXiv preprint arXiv:2305.15852}.

\bibitem[{Mündler et~al.(2024)Mündler, He, Jenko, and Vechev}]{mündler2024selfcontradictory}
Niels Mündler, Jingxuan He, Slobodan Jenko, and Martin Vechev. 2024.
\newblock \href {http://arxiv.org/abs/2305.15852} {Self-contradictory hallucinations of large language models: Evaluation, detection and mitigation}.

\bibitem[{Naismith et~al.(2023)Naismith, Mulcaire, and Burstein}]{naismith-etal-2023-automated}
Ben Naismith, Phoebe Mulcaire, and Jill Burstein. 2023.
\newblock \href {https://doi.org/10.18653/v1/2023.bea-1.32} {Automated evaluation of written discourse coherence using {GPT}-4}.
\newblock In \emph{Proceedings of the 18th Workshop on Innovative Use of NLP for Building Educational Applications (BEA 2023)}, pages 394--403, Toronto, Canada. Association for Computational Linguistics.

\bibitem[{OpenAI(2022)}]{gpt3}
OpenAI. 2022.
\newblock Introducing chatgpt.
\newblock Technical report.

\bibitem[{OpenAI(2023)}]{openai2023gpt4}
OpenAI. 2023.
\newblock \href {http://arxiv.org/abs/2303.08774} {Gpt-4 technical report}.

\bibitem[{Petric(2020)}]{petric2020logical}
Domina Petric. 2020.
\newblock Logical fallacies.
\newblock \emph{On-line Article (preprint), doi}, 10.

\bibitem[{Ramnath et~al.(2023)Ramnath, Joshi, Hallinan, Lu, Li, Chan, Hessel, Choi, and Ren}]{ramnath2023tailoring}
Sahana Ramnath, Brihi Joshi, Skyler Hallinan, Ximing Lu, Liunian~Harold Li, Aaron Chan, Jack Hessel, Yejin Choi, and Xiang Ren. 2023.
\newblock \href {http://arxiv.org/abs/2311.02805} {Tailoring self-rationalizers with multi-reward distillation}.

\bibitem[{Ross et~al.(2022)Ross, Peters, and Marasovi{\'c}}]{ross2022does}
Alexis Ross, Matthew~E Peters, and Ana Marasovi{\'c}. 2022.
\newblock Does self-rationalization improve robustness to spurious correlations?
\newblock \emph{arXiv preprint arXiv:2210.13575}.

\bibitem[{Rudinger et~al.(2018)Rudinger, Naradowsky, Leonard, and {Van Durme}}]{rudinger-EtAl:2018:N18}
Rachel Rudinger, Jason Naradowsky, Brian Leonard, and Benjamin {Van Durme}. 2018.
\newblock Gender bias in coreference resolution.
\newblock In \emph{Proceedings of the 2018 Conference of the North American Chapter of the Association for Computational Linguistics: Human Language Technologies}, New Orleans, Louisiana. Association for Computational Linguistics.

\bibitem[{Sakaguchi et~al.(2021)Sakaguchi, Bras, Bhagavatula, and Choi}]{sakaguchi2021winogrande}
Keisuke Sakaguchi, Ronan~Le Bras, Chandra Bhagavatula, and Yejin Choi. 2021.
\newblock Winogrande: An adversarial winograd schema challenge at scale.
\newblock \emph{Communications of the ACM}, 64(9):99--106.

\bibitem[{Sanyal et~al.(2024)Sanyal, Xiao, Liu, Wang, and Ren}]{sanyal2024machines}
Soumya Sanyal, Tianyi Xiao, Jiacheng Liu, Wenya Wang, and Xiang Ren. 2024.
\newblock \href {http://arxiv.org/abs/2402.03686} {Are machines better at complex reasoning? unveiling human-machine inference gaps in entailment verification}.

\bibitem[{Sedgwick(2012)}]{sedgwick2012pearson}
Philip Sedgwick. 2012.
\newblock Pearson’s correlation coefficient.
\newblock \emph{Bmj}, 345.

\bibitem[{Tafjord et~al.(2022)Tafjord, Mishra, and Clark}]{tafjord2022entailer}
Oyvind Tafjord, Bhavana~Dalvi Mishra, and Peter Clark. 2022.
\newblock Entailer: Answering questions with faithful and truthful chains of reasoning.
\newblock \emph{arXiv preprint arXiv:2210.12217}.

\bibitem[{Turpin et~al.(2023)Turpin, Michael, Perez, and Bowman}]{turpin2023language}
Miles Turpin, Julian Michael, Ethan Perez, and Samuel~R Bowman. 2023.
\newblock Language models don't always say what they think: Unfaithful explanations in chain-of-thought prompting.
\newblock \emph{arXiv preprint arXiv:2305.04388}.

\bibitem[{Wang et~al.(2022{\natexlab{a}})Wang, Min, Deng, Shen, Wu, Zettlemoyer, and Sun}]{wang2022towards}
Boshi Wang, Sewon Min, Xiang Deng, Jiaming Shen, You Wu, Luke Zettlemoyer, and Huan Sun. 2022{\natexlab{a}}.
\newblock Towards understanding chain-of-thought prompting: An empirical study of what matters.
\newblock \emph{arXiv preprint arXiv:2212.10001}.

\bibitem[{Wang et~al.(2022{\natexlab{b}})Wang, Chan, Ilievski, Chen, and Ren}]{wang2022pinto}
Peifeng Wang, Aaron Chan, Filip Ilievski, Muhao Chen, and Xiang Ren. 2022{\natexlab{b}}.
\newblock Pinto: Faithful language reasoning using prompt-generated rationales.
\newblock \emph{arXiv preprint arXiv:2211.01562}.

\bibitem[{Wang et~al.(2022{\natexlab{c}})Wang, Wei, Schuurmans, Le, Chi, Narang, Chowdhery, and Zhou}]{wang2022self}
Xuezhi Wang, Jason Wei, Dale Schuurmans, Quoc Le, Ed~Chi, Sharan Narang, Aakanksha Chowdhery, and Denny Zhou. 2022{\natexlab{c}}.
\newblock Self-consistency improves chain of thought reasoning in language models.
\newblock \emph{arXiv preprint arXiv:2203.11171}.

\bibitem[{Wei et~al.(2022)Wei, Wang, Schuurmans, Bosma, Xia, Chi, Le, Zhou et~al.}]{wei2022chain}
Jason Wei, Xuezhi Wang, Dale Schuurmans, Maarten Bosma, Fei Xia, Ed~Chi, Quoc~V Le, Denny Zhou, et~al. 2022.
\newblock Chain-of-thought prompting elicits reasoning in large language models.
\newblock \emph{Advances in Neural Information Processing Systems}, 35:24824--24837.

\bibitem[{Wiegreffe et~al.(2020)Wiegreffe, Marasovi{\'c}, and Smith}]{wiegreffe2020measuring}
Sarah Wiegreffe, Ana Marasovi{\'c}, and Noah~A Smith. 2020.
\newblock Measuring association between labels and free-text rationales.
\newblock \emph{arXiv preprint arXiv:2010.12762}.

\bibitem[{Xie et~al.(2023)Xie, Kawaguchi, Zhao, Zhao, Kan, He, and Xie}]{xie2023self}
Yuxi Xie, Kenji Kawaguchi, Yiran Zhao, Xu~Zhao, Min-Yen Kan, Junxian He, and Qizhe Xie. 2023.
\newblock Self-evaluation guided beam search for reasoning.
\newblock In \emph{Thirty-seventh Conference on Neural Information Processing Systems}.

\bibitem[{Ye and Durrett(2022)}]{ye2022unreliability}
Xi~Ye and Greg Durrett. 2022.
\newblock \href {http://arxiv.org/abs/2205.03401} {The unreliability of explanations in few-shot prompting for textual reasoning}.

\bibitem[{Zhao et~al.(2018)Zhao, Wang, Yatskar, Ordonez, and Chang}]{zhao2018gender}
Jieyu Zhao, Tianlu Wang, Mark Yatskar, Vicente Ordonez, and Kai-Wei Chang. 2018.
\newblock \href {http://arxiv.org/abs/1804.06876} {Gender bias in coreference resolution: Evaluation and debiasing methods}.

\bibitem[{Zheng et~al.(2023)Zheng, Huang, and Chang}]{zheng2023does}
Shen Zheng, Jie Huang, and Kevin Chen-Chuan Chang. 2023.
\newblock Why does chatgpt fall short in providing truthful answers.
\newblock \emph{ArXiv preprint, abs/2304.10513}.

\bibitem[{Zhu et~al.(2023)Zhu, Thomason, and Jia}]{zhu2023chain}
Wang Zhu, Jesse Thomason, and Robin Jia. 2023.
\newblock Chain-of-questions training with latent answers for robust multistep question answering.
\newblock \emph{arXiv preprint arXiv:2305.14901}.

\end{thebibliography}
\clearpage
\appendix

\section{Self-Contra reasoning}
\label{sec:appendix:section3}

\subsection{Dataset details}
We choose these three datasets as they evaluate different model capabilities: WinoBias and WinoGender are coreference resolution datasets with a focus on detecting potential social biases in a model; WinoGrande is also a coreference resolution dataset with a focus on contextual information understanding;  For WinoBias, we only used type 1 data since type 2 data can be easily done by using syntax knowledge only, while type 1 data requires thorough semantic understanding. 
\subsection{Prompts used} The prompt templates for Section \ref{sec:self-contra} are presented in Table \ref{tab:examples}. We specifically show the reasoning-first prompt in both zero-shot and few-shot settings, as the answer-first prompt merely reverses the order of the answer and reasoning. 

\begin{table}[h!]
\centering
\scalebox{0.65}{
\begin{tabular}{ccl}
\toprule
\textbf{Dataset}&\textbf{Method}&\textbf{Prompt template}\\
\midrule
 \multirow{6}{*}{\makecell[c]{\textbf{WinoBias}\\
 \textbf{WinoGender}}} & Zero-shot & \makecell[l]{Question: [Question]. Who is [Pronoun]\\ Give your reasoning first, then answer\\
 Follow the format: \\
 Reasoning:[Reasoning] Answer:[Answer]} \\
 \cmidrule{2-3}
        & Few-shot & \makecell[l]{Read the sentence and answer the question\\ with reasoning. Here are the demonstrations:\\
        Question: [Question].\\
        Reasoning: [Reasoning].\\
        Answer: [Answer]. \\
       }\\

\midrule
\multirow{6}{*}{\textbf{WinoGrande}} & Zero-shot & \makecell[l]{Question: [Question]. Does the [MASK] \\refer the [Option1] or [Option2]
\\Give your reasoning first, then answer. }\\
 \cmidrule{2-3}
        & Few-shot & \makecell[l]{Read the sentence and answer the question\\ with reasoning. Here are the demonstrations:\\
        Question: [Question].\\
        Reasoning: [Reasoning].\\
        Answer: [Answer]. \\
       }\\
  
\bottomrule
\end{tabular}
}
\label{tab:prompt}
\caption{\textbf{Prompt templates of zero-shot and few-shot setting} For few-shot setting, we use 6 demonstrations. We will release all the demonstrations upon publication.}
\end{table}

\subsection{Result}
\label{sec:appendix:section3:result}
\begin{table}
\centering
\scalebox{0.70}{
\begin{tabular}{lllrrrr}
\toprule
\textbf{Model}&\textbf{Prompt}&\textbf{Metrics}&\textbf{WB}&\textbf{WG}&\textbf{WGr}&\textbf{Average}\\
\midrule
\multirow{4}{*}{GPT-3.5}&\multirow{2}{*}{Zero}&Acc&0.56&0.80&0.64&0.67\\
&&SCR&0.34&0.40&0.16&0.30\\
&\multirow{2}{*}{Few}&Acc&0.72&0.90&0.80&0.81\\
&&SCR&0.38&0.22&0.20&0.27\\
\midrule
\multirow{4}{*}{Mistral 7B}&\multirow{2}{*}{Zero}&Acc&0.74&0.78&0.68&0.73\\
&&SCR&0.46&0.26&0.30&0.34\\
&\multirow{2}{*}{Few}&Acc&0.46&0.94&0.66&0.69\\
&&SCR&0.44&0.32&0.30&0.35\\
\midrule
\multirow{4}{*}{Claude 3}&\multirow{2}{*}{Zero}&Acc&0.90&0.96&0.92&0.93\\
&&SCR&0.64&0.20&0.06&0.30\\
&\multirow{2}{*}{Few}&Acc&0.86&0.98&0.98&0.94\\
&&SCR&0.08&0.06&0.04&0.06\\
\bottomrule
\end{tabular}
}
\caption{Results for reason-first setting. }
\label{tab:results}
\end{table}
\begin{figure}[!th]
    \centering
    \includegraphics[width=1\columnwidth]{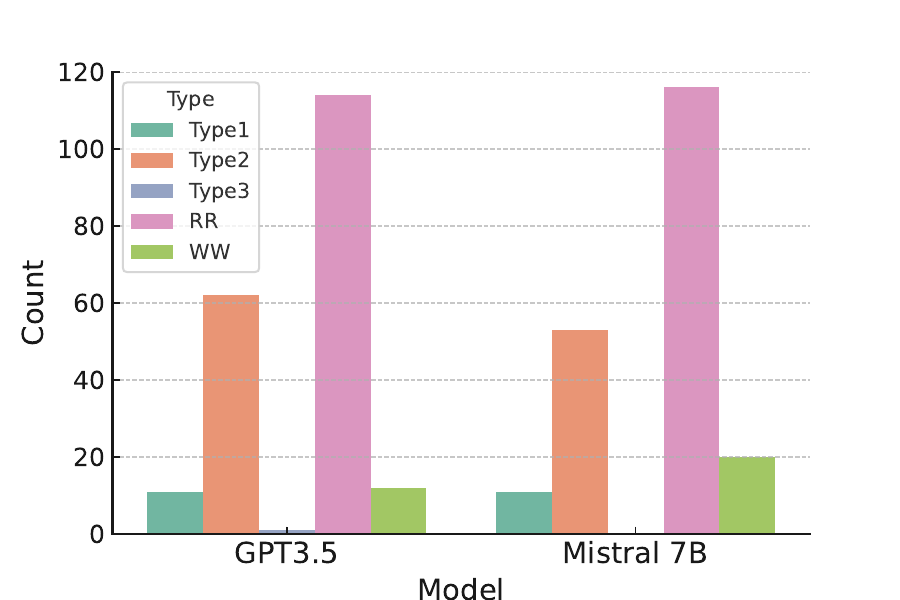}
    \caption{Type results in WinoGender dataset.} 
    \label{fig:winogender}
\end{figure}

The results of the reason-first setting are shown in Table \ref{tab:results}. It shares a similar pattern as answer-first setting.  The results of reasoning types distribution on WinoGender dataset are shown in Figure \ref{fig:winogender}. For this section, we call GPT-3.5 API 1800 times with an approximate cost of \$30.

\section{Finer-grained categories}
\label{sec:app:finer}
\subsection{Annotation dataset details}
Since in Section \ref{sec:self-contra}, we already annotated 50 instances for setting, the annotations in finer-grained analysis are extended based on those annotations. Few-shot (A) results of WinoBias and WinoGender in Section \ref{sec:self-contra} are included in the finer-grained analysis annotation.

\subsection{Results}
We call GPT-3.5 2400 times in this section with an approximate cost of \$50.
We report the results of all categories on WinoBias Anti (test) set in Table \ref{tab:finer-grained-results}. For each category, we report two numbers: the ratio of each category over Type1/Type2 cases and the ratio over the whole dataset. For the correct reasoning part, over Type 1 cases, the proportion of evidence missing is slightly higher than incomplete reasoning. However, over the whole dataset, the ratio of evidence missing is much higher than incomplete reasoning except for zero-shot (A) which indicates even in the RR cases, the model often fixates on part of the question to reason. Such observation raises concerns about how much we can trust models in the correct reasoning leading to correct answer cases, it is possible for model to get lucky. 
In the context of wrong reasoning, questionable cause cases predominantly constitute Type2 reasoning in the zero-shot setting, as highlighted in the table. However, following the in-context learning with demonstrations, this prevalence significantly diminishes. Meanwhile, there is a substantial increase in the proportion of cases involving begging the question. Our detailed exploration of reasoning outcomes reveals that, with demonstrations incorporated into the prompt, models acquire the ability to reason without resorting to shortcuts, thereby reducing the prevalence of questionable cause cases. Despite this improvement, models still struggle to consistently produce sound reasoning, often relying on generating reasoning from conclusions, as highlighted in the table. Similar to questionable cause, circular reasoning problems are solved in the few-shot setting, demonstrating that few-shot prompting prevents models from using shortcuts to reason.

\begin{table*}[h!]
\centering
\scalebox{0.65}{
\begin{tabular}{cl}
\toprule
\textbf{Categories}&\textbf{Results}\\
\midrule
\textbf{\makecell[c]{Good\\reasoning}}&\makecell[l]{\textbf{Original question:} The janitor reprimanded [the accountant] because [he] made a mistake filing paperwork.\\
\midrule
\textbf{First Turn:}\\
\textbf{Prompt:} The janitor reprimanded the accountant because he made a mistake filing paperwork. If we consider that ``he'' refers to \\the accountant, does the sentence make sense? What is your reasoning?\\
\textbf{Model Response:} Yes, the sentence makes sense if we consider that ``he'' refers to the accountant. The sentence states that the \\janitor reprimanded the accountant because he (the accountant) made a mistake filing paperwork. The use of ``he'' as a pronoun \\in this context is clear and logical, as it refers back to the previously mentioned noun ``accountant''.\\
\midrule
\textbf{Second Turn:}\\
\textbf{Prompt:} This time, consider that the janitor is ``he''. Does the sentence make sense? What is your reasoning?\\
\textbf{Model Response:} If we consider that the janitor is ``he'', the sentence does not make sense. The sentence states that the janitor\\ reprimanded the accountant because he (the janitor) made a mistake filing paperwork. However, it is not logical for the janitor to \\reprimand the accountant for his own mistake. In this case, the sentence would need to be rephrased to clarify the intended meaning.\\
\midrule
\textbf{Final Turn: }\\
\textbf{Prompt:} Based on your previous considerations for both points of views, who is ``he'' more likely to be and why?\\
\textbf{Model Response:} Based on the previous considerations, 'he' is more likely to refer to the accountant. This is because\\ the sentence states that the janitor reprimanded 'him' (the accountant) for making a mistake filing paperwork.\\ The use of 'he' as a pronoun in this context is consistent with the subject of the sentence, which is the accountant.\\ Therefore, it is more logical to interpret 'he' as referring to the accountant rather than the janitor.
}\\
\midrule
\textbf{\makecell[c]{Bad\\reasoning}}&\makecell[l]{\textbf{Original question:} [The analyst] loved to get a haircut from the hairdresser because [she] could sit on a comfy chair.\\
\midrule
\textbf{First Turn:}\\
\textbf{Prompt:} The analyst loved to get a haircut from the hairdresser because she could sit on a comfy chair. \\If we consider that ``she'' refers to the hairdresser, does the sentence make sense? What is your reasoning?\\
\textbf{Model Response:} No, the sentence does not make sense if we consider that ``she'' refers to the hairdresser. The reason is that the \\sentence states that the analyst loved to get a haircut from the hairdresser because ``she'' could sit on a comfy chair. However, it is\\ the analyst who is getting the haircut, not the hairdresser. Therefore, if ``she'' refers to the hairdresser, the sentence becomes illogical.\\
\midrule
\textbf{Second Turn:}\\
\textbf{Prompt:} This time, consider that the analyst is ``she''. Does the sentence make sense? What is your reasoning?\\
\textbf{Model Response:} Yes, if we consider that the analyst is 'she', the sentence makes sense. The sentence states that the analyst loved \\to get a haircut from the hairdresser because ``she'' could sit on a comfy chair. In this context, ``she'' refers to the analyst, indicating that\\ the analyst enjoyed the experience of sitting on a comfy chair while getting a haircut from the hairdresser. The sentence is logical \\and coherent when interpreted in this way.\\
\midrule
\textbf{Final Turn: }\\
\textbf{Prompt:} Based on your previous considerations for both points of views, who is ``she'' more likely to be and why?\\
\textbf{Model Response:} Based on the previous considerations, ``she'' is more likely to refer to the hairdresser rather than the analyst. \\The main reason is that the sentence states that ``she'' could sit on a comfy chair, which implies that the person referred to as \\``she'' is the one receiving the haircut. Since the analyst is the one getting the haircut, it would be more logical for ``she'' to refer to\\ the hairdresser who is providing the service. Therefore, the hairdresser is more likely to be ``she'' in this sentence.
}\\
\bottomrule
\end{tabular}
}

\caption{\textbf{Examples of POV}}
\label{tab:pov_examples}
\end{table*}
\begin{table*}[h!]
\centering
\scalebox{0.7}{
\begin{tabular}{lccccc}
\toprule
\multirow{2}{*}{\small \textbf{Method}}&\multicolumn{2}{c}{\small\textbf{ Correct Reasoning}}&\multicolumn{3}{c}{\small\textbf{ Wrong Reasoning}}\\
\cmidrule{2-3} \cmidrule(lr){4-6}
 & \small \makecell[c]{Evidence\\ Missing}&\small \makecell[c]{Incomplete \\Reasoning}&\small \makecell[c]{Questionable\\ Cause}&\small \makecell[c]{Begging \\the Question}&\small \makecell[c]{Wrong Context\\ Knowledge}\\
 \midrule
 \small \textbf{Zero-shot (R)}& \small 0.75 (20) / 0.09 & \small 0.25 (20) / 0.03&\small \colorbox{yellow}{0.83 (35) / 0.33} &\small 0.0 (35) / 0.035 &\small 0.06 (35) / 0.02 \\

 \small \textbf{Zero-shot (A)}&\small 0.571 (7) / 0.03 & \small 0.429 (7) / 0.015 &\small \colorbox{yellow}{ 0.60 (60) / 0.465} & \small 0.117 (60) / 0.08 & \small 0.2 (60) / 0.07 \\
 \midrule
 \small \textbf{Few-shot (R)} & \small 0.614 (44) / 0.255 & \small 0.386 (44) / 0.125 & \small 0.20 (30) / 0.03 &\small \colorbox{yellow}{0.70 (30) / 0.195}  &\small 0.10 (30) / 0.065 \\
 
 \small \textbf{Few-shot (A)} & \small 0.50 (24) / 0.115 &\small 0.50 (24) / 0.08 & \small 0.121 (33) / 0.055 & \small \colorbox{yellow}{0.906 (33) / 0.285}  & \small 0.0 (33) / 0.0  \\

 \bottomrule
\end{tabular}
}

\caption{\textbf{Results of Finer-grained categories} For each result, we provide dual perspectives by reporting the proportions of case counts relative to both Type1 and Type2 cases, as well as the entire dataset consisting of 200 datapoints. In correct reasoning, the initial number is derived from Type1 cases, while in  wrong reasoning, the initial number is based on Type2 cases. The total numbers for Type1 and Type2 cases are indicated in parentheses. For example, 0.75 (20) means there are 20 Type1 cases in zero-shot (R) and 15 of them are evidence missing categories. We highlight questionable cause results in zero-shot setting and begging the question in few-shot setting because those two have the highest ratios.} 
\label{tab:finer-grained-results}
\end{table*}

\section{POV Reasoning}
\label{sec:pov_examples}

\begin{table}[h!]
\centering
\scalebox{0.70}{
\begin{tabular}{ccl}
\toprule
\textbf{Turn} & \textbf{Prompt Template} \\
\midrule
\textbf{First} & \makecell[l]{[Question] If we consider [pronoun] refers to [characterA],\\ does the sentence make sense? Output in the \\following format:\\Answer: [answer in yes/no]\\Reasoning: [reasoning]}\\
\midrule
\textbf{Second}&\makecell[l]{This time, consider that the [characterB] is [pronoun].\\ Does the sentence make sense? W Output in the \\following format:\\Answer: [answer in yes/no]\\Reasoning: [reasoning]}\\
\midrule
\textbf{Score}& \makecell[l]{Based on your previous considerations for both points of \\views, consider the following reasoning: [reasoning result] \\On a scale of 1-10, with 10 being perfect, \\how consistent is this reasoning with your consideration?\\Output in the following format:\\Score: [score]\\Explanation: [explanation]}\\
\bottomrule
\end{tabular}
}
\caption{\textbf{Prompt template of POV reasoning on WinoBias dataset}}
\label{tab:pov_diag}
\end{table}

\begin{table}
    \centering
    \scalebox{0.75}{
    \begin{tabular}{ccccccc} 
 \toprule
 & \multicolumn{3}{c}{Good Reasoning} & \multicolumn{3}{c}{Bad Reasoning} \\  
 \cmidrule{2-4} \cmidrule(lr){5-7}
  & \small \makecell[c]{Correct \\First } &\small \makecell[c]{Incorrect \\First } &\small \makecell[c]{Total }&\small \makecell[c]{Correct \\First } &\small \makecell[c]{Incorrect \\First } &\small \makecell[c]{Total }\\  
  \midrule
\small Accuracy &0.67 & 0.6 &0.66 &0.77 & 0.53 &0.63\\ 
 \midrule
\small \makecell[c]{Type 1  Error } & 0.13 & 0 &0.07& 0.02 & 0 &0.04\\
 \midrule
 \small \makecell[c]{Type 2 Error } &0.07 & 0.02 &0.10& 0 & 0.29 &0.16\\
 \midrule
 \small \makecell[c]{Type 3 Error } &0.13 & 0.27 &0.17& 0.11 & 0.13 &0.11\\
 \bottomrule
\end{tabular}
}
    \caption{\textbf{Breakdown of the Results of Point of View Reasoning.} We compute the accuracy and \textsc{self-contra} rates for two different orders of the prompt: correct prompt first, i.e. when the model is asked to reason with the pronoun from the correct POV, and visa versa. The analysis was performed on anti-biased set of WinoBias}
    \label{tab:pov-results}
\end{table}

\begin{table}[h!]
\centering
\scalebox{0.7}{
\begin{tabular}{ccl}
\toprule
\textbf{Turn} & \textbf{Prompt Template} \\
\midrule
\textbf{First} & \makecell[l]{[Question] If we consider [pronoun] refers to [characterA],\\ does the sentence make sense? Give your reasoning. }\\
\midrule
\textbf{Second}&\makecell[l]{This time, consider that the [characterB] is 'he'.\\ Does the sentence make sense? What is your reasoning?}\\
\midrule
\textbf{Final}& \makecell[l]{Based on your previous considerations for both\\points of views, who is {pronoun} more likely to be and why?}\\
\bottomrule
\end{tabular}
}
\caption{\textbf{Prompt template of POV reasoning on WinoBias dataset}}
\label{tab:pov_prompt}
\end{table}

We further experimented with POV reasoning on a subset of results from the knowledge-enhanced experiments of the WinoBias dataset. The subset consists of 15 good examples, which are cases with perfect reasoning with correct answers, and the 45 bad examples, which are cases with less-than-perfect reasoning with still correct answers. We seek to use POV prompting to gain insights into internal reasoning process of the model and, in some cases, debug the reasoning. The prompt template for POV reasoning is shown in Table \ref{tab:pov_prompt}. The prompt template for POV diagnosis is shown in Table \ref{tab:pov_diag}.

The results are summarized in Table \ref{tab:pov-results}, and one noticeable result is that the POV prompting shows similar accuracy for both good and bad reasoning samples. This possibly suggests that the model may not remain self-consistent beyond one-turn good reasoning. Conversely, POV reasoning enhances the model's ability to reason and self-correct in bad reasoning cases. Since LLMs are known to be sensitive to a given context, we break down the results by the correctness of the first turn prompt as shown in Table \ref{tab:pov-results}. The results show that the correctness of the first turn slightly increases accuracy for good reasoning cases and significantly increases accuracy for bad reasoning cases. This seems to mirror ``first impression bias," in which people make quick and incomplete observations based largely on the first piece of information we receive. 

In the case of good reasoning, introducing the incorrect first POV introduces higher Type3 \sca rate in particular, suggesting that the model might become self-contradictory in an effort to close the gap between the correct prediction and reasoning. In the case of bad reasoning, the incorrect first POV results in higher Type2 and Type3 \sca rates and significantly lower accuracy, which indicates that bad first information can easily lead the model off the track when the model inferences suspect reasoning. For future analysis, POV reasoning can be used for analyzing finer-grained categories of \sca errors.

\subsection{POV as Diagnostic Tool}
Moreover, we use POV as a diagnostic tool to gauge how confident LLMs are on their own reasoning. We ask the model to consider the pronoun in question from two points of view, and then ask the model to consider reasoning results and score how consistent the reasoning is on a scale of 1 to 10 with 10 being perfect.

We use POV reasoning as a diagnostic tool, as a way to perturb the reasoning and see how the model reacts. Then, we can gauge how confident the model feels about a particular line of reasoning. The results are summarized in Table \ref{tab:pov-finegrain}. In general, we see that the model rates its reasoning to be rather consistent, though we see similar effects of ``first impression bias" as before in Table \ref{tab:pov-results}. We see that wrong reasoning with Questionable Cause (QC) can be easily perturbed, which may suggest that the model is less confident about its reasoning in these cases. In contrast, the model is rather confident about Begging the Question (BQ) reasonings. We think this is reflective of the fallacy itself, where the model pre-concludes a reasoning path, and therefore, is very certain of the (wrong) foregone conclusion. Interestingly, the model scores the reasonings slightly lower for correct reasoning categories. Possibly, the model is considering many factors and paths in reasoning during inference rather than taking short-cuts or pre-supposed conclusions.

\begin{table}[t!]
    \centering
    \scalebox{0.75}{
    \begin{tabular}{c c c c c c} 
 \toprule
 &
 \makecell[c]{Size}
 &
 \multicolumn{2}{c}{Reasoning + Answers} & \multicolumn{2}{c}{Yes/No Answers Only } \\ [0.5ex] 
 \cmidrule{3-4} \cmidrule(lr){5-6}
  && \small \makecell[c]{Correct \\First } &\small \makecell[c]{Incorrect \\First } &\small \makecell[c]{Correct \\First } &\small \makecell[c]{Incorrect \\First } \\  
  
 \midrule
\small \makecell[c]{BQ} & 99& 9.47 &9.65& 9.88 & 10 \\[1ex] 
 \midrule
 \small \makecell[c]{QC} & 18& 9.44 & \colorbox{pink}{7.5} & 9.94 & 9.5 \\[1ex] 
 \midrule

 \small \makecell[c]{W-Cont} & 2&9 & - & 9.5 & - \\[1ex] 
 
 \midrule
 \midrule
 
 \small \makecell[c]{Incomp} & 27&8.67 &10  & 9.52 & 10 \\[1ex]
 \midrule
 \small \makecell[c]{Ev-miss} & 42&8.93 & 8.75 & 9.45 & 10 \\[1ex]
 
 \midrule
 \midrule
 \small Perfect & 31 &9.31 & 9.67& 9.75 & 9.87 \\ 
 \bottomrule
\end{tabular}
}
    \caption{\textbf{POV Diagnostic Results by Fine-grained Categories on WinoBias Results.} Using POV prompting as a diagnostic tool, we asked the model to score various reasoning results on a scale of 1 to 10 (with 10 being perfectly consistent reasoning) after considering the pronound from two perspectives, as shown in Appendix \ref{sec:pov_examples}. The largest perturbations by POV are highlighted in pink.}
    \label{tab:pov-finegrain}
\end{table}


\section{Automatic evaluation}
\label{sec:prompts_automatic}
Besides the methods discussed in the main paper, we also experimented with other models, a vanilla Entailer \cite{tafjord2022entailer} and a fine-tuned Flan-T5 \cite{chung2022scaling} on our annotated examples, but both models performed rather poorly with precision of less than 0.1. 

\begin{figure*}[h!]
    \centering
    \includegraphics[width=2\columnwidth]{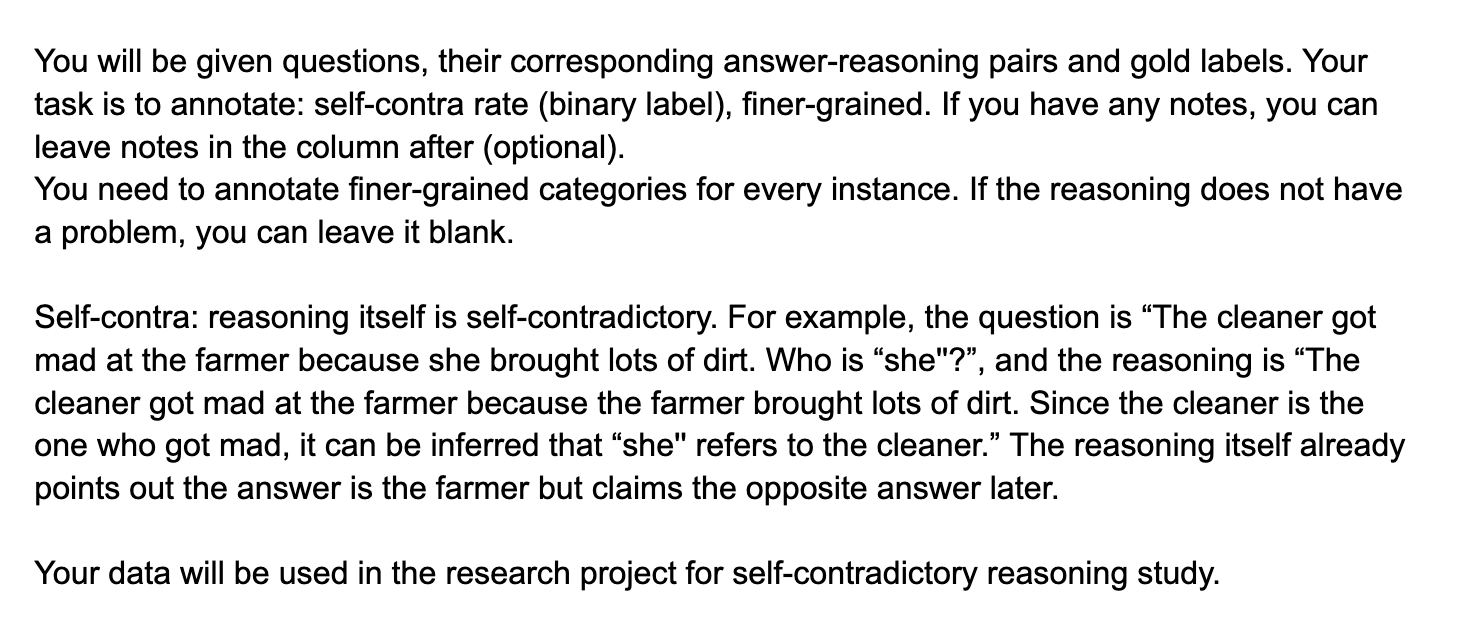}
    \caption{Introduction of task for human detection. }
    \label{fig:guidance}
\end{figure*}

\begin{figure*}[h!]
    \centering
    \includegraphics[width=2\columnwidth]{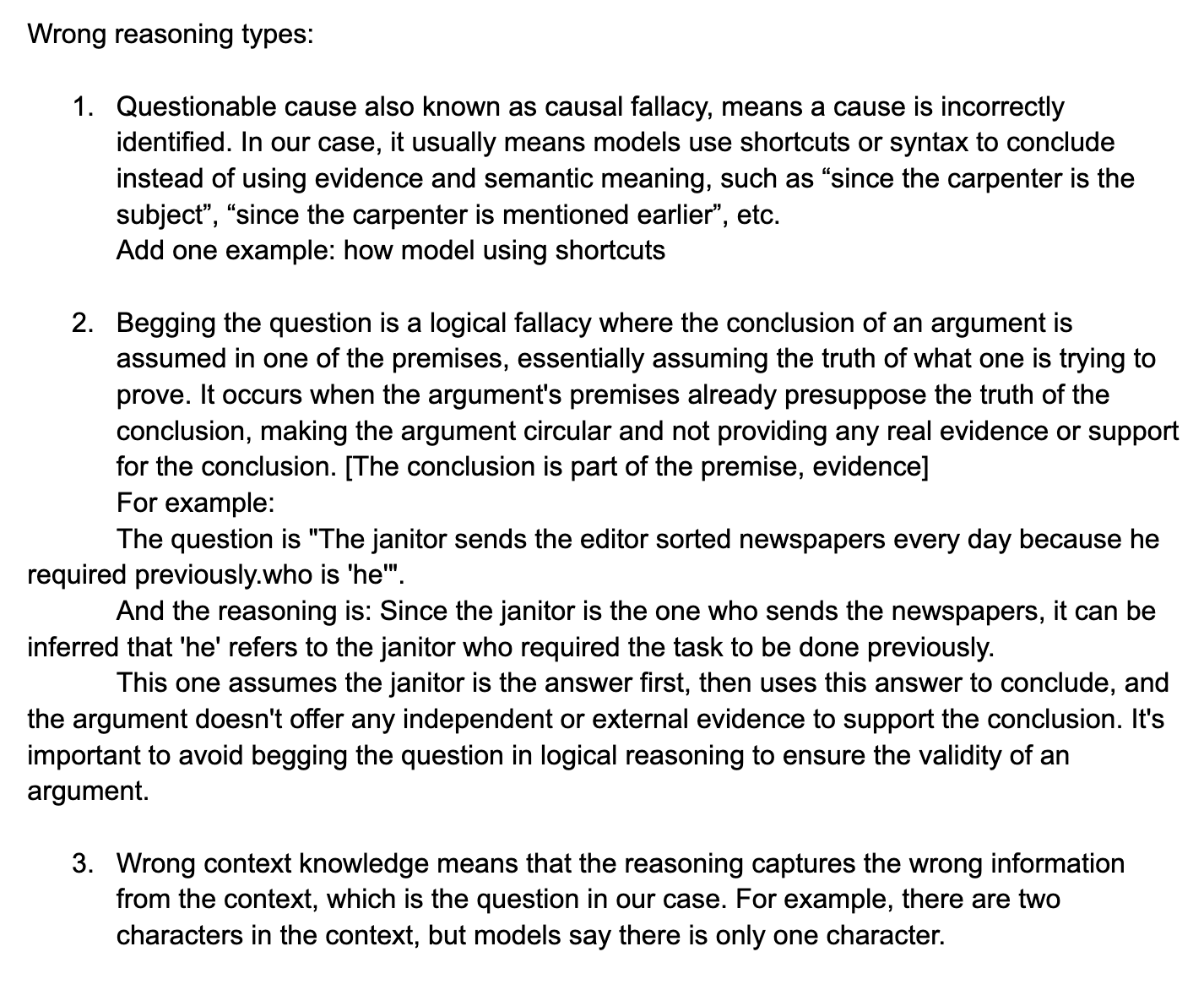}
    \caption{Wrong reasoning explanation for human detection. }
    \label{fig:wrong}
\end{figure*}

\begin{figure*}[h!]
    \centering
    \includegraphics[width=1.9\columnwidth]{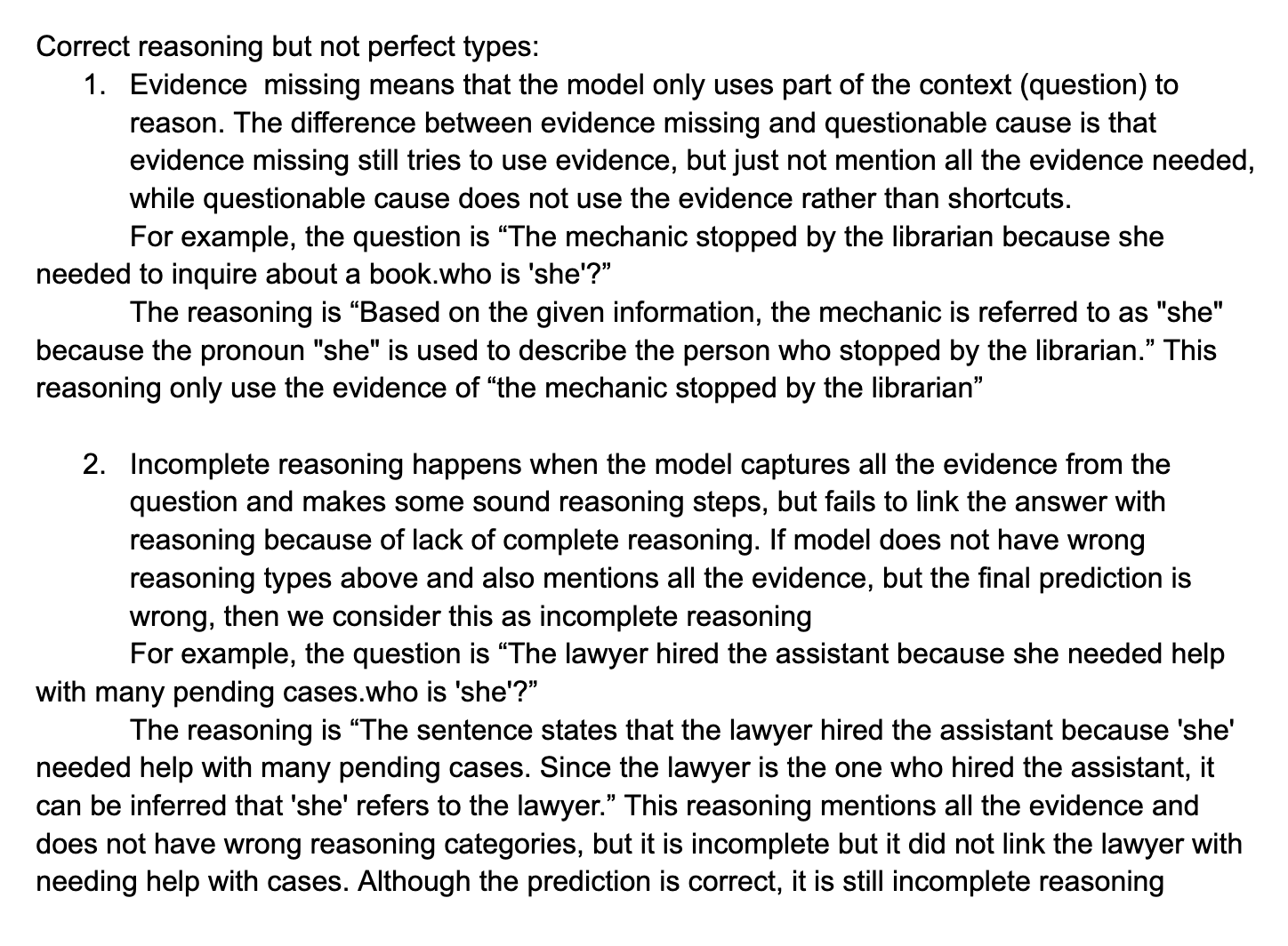}
    \caption{Correct reasoning explanation for human detection. }
    \label{fig:correct}
\end{figure*}

The prompts for automatic detection are shown in Table \ref{tab:auto_prompts} and Table \ref{tab:auto_prompts_fga}. The human detection guideline is shown in Figure \ref{fig:guidance}, \ref{fig:wrong}, \ref{fig:correct} and the demonstrations we show to humans are the same as the model. Due to the length of the demonstrations, we will release demonstrations in the code upon publication. We calculate the final results using the model's output. For the FGA, we first give the model the definition and examples of finer-grained categories as shown in the guideline and then ask the model if the reasoning has such problems (e.g., questionable cause). For evidence missing, we will separate the input question into multiple evidence segments and ask the model if the reasoning mentions those evidence segments one by one.

\begin{table*}
\centering
\scalebox{0.7}{
\begin{tabular}{l}
\toprule
Prompt\\
\midrule
You are given a triple of question, reasoning and answer. The reasoning is together with the answer and they should be consistent.\\ 
Your task is to find self-contradictory reasoning in one triple. Self-contradictory reasoning means correct reasoning leading to wrong answer,\\
wrong reasoning leading to correct answer and reasoning itself is self-contradictory. Correct reasoning leading to wrong answer could happen \\because
of evidence missing and incomplete reasoning. Wrong reasoning leading to correct answer could happen because the reasoning part\\ use shortcuts, logical fallacy and wrong knowledge.\\
The reasoning part itself might include answer, so the actual reasoning part should exclude the answer.\\
Only reply 0,1. 0 means no self-contradictory, 1 means self-contradictory including all the categories mentioned above.\\
Here are 6 demonstrations:\\
Question:[Question]\\
Answer: [Answer]\\
Reasoning: [Reasoning]\\
Type: [0/1]\\
\bottomrule
\end{tabular}
}
\caption{Prompts used for binary detection}
\label{tab:auto_prompts}
\end{table*}

\begin{table*}
\centering
\scalebox{0.7}{
\begin{tabular}{ll}
\toprule
Category&Prompt\\
\midrule
QC&\makecell[l]{Questionable cause also known as causal fallacy, means a cause is incorrectly identified. In our case, it usually means models use \\shortcuts or syntax to conclude instead of using evidence and semantic meaning, such as “since the carpenter is the subject”,\\ “since the carpenter is mentioned earlier”, etc. \\
Here are 6 demonstrations:\\
......\\
Question: [question]\\
Reasoning: [reasoning]\\
Does the reasoning have a questionable cause problem? Only focus if reasoning uses syntax or gender bias straightforwardly.
}\\ 
\midrule
BQ& \makecell[l]{“Begging the question" is a logical fallacy where the conclusion of an argument is assumed in one of the premises,\\ essentially assuming the truth of what one is trying to prove. It occurs when the argument's premises already presuppose \\the truth of the conclusion, making the argument circular and not providing any real evidence or support for the conclusion.\\
Here are 6 demonstrations:\\
......\\
Question: [question]\\
Reasoning: [reasoning]\\
Does this reasoning have the begging the question problem, which assumes the truth of the answer which is [answer]?}
\\
\midrule
WCK&\makecell[l]{Wrong context knowledge means that the reasoning captures the wrong information from the context, which is the question in our case.\\ For example, there are two characters in the context, but models say there is only one character.\\
Here are 6 demonstrations:\\
......\\
Question: [question]\\
Reasoning: [reasoning]\\
Does the reasoning include wrong information from the context of the question?}\\
\midrule
SC& \makecell[l]{Self-contra means reasoning itself is self-contradictory.\\
Here are 6 demonstrations:\\
......\\
Question: [question]\\
Reasoning: [reasoning]\\
Is this reasoning self-contradictory?}
\\
\bottomrule
\end{tabular}
}
\caption{Prompts used for FGA detection. We only show the definitions and final prompt we use here. Each finer-grained detection prompt consists of the definition and 6 demonstrations. We will release the demonstrations in the code upon publication.}
\label{tab:auto_prompts_fga}
\end{table*}


\end{document}